\documentclass[journal,12pt,onecolumn]{IEEEtran}
\usepackage{graphicx}
\usepackage{xcolor}
\usepackage{subfigure}
\usepackage{listings}
\usepackage{verbatim}

\begin{document}

\title{Clustering Time Series Data through Autoencoder-based Deep Learning Models}

%Clustering Financial Time Series Data Through  Autoencoder-Based Deep Learning
%
%\titlerunning{Clustering Financial Time Series Through Deep Learning}

\author{Neda~Tavakoli,
        Sima~Siami-Namini, 
        Mahdi~Adl Khanghah,
        Fahimeh~Mirza Soltani,
        and~Akbar~Siami~Namin%*\thanks{* Corresponding author.}
        % <-this % stops a space
\thanks{N. Tavakoli is with the Department
of Computer Science, Georgia Institute of Technology, Atlanta,
GA, 30332 USA e-mail: neda.tavakoli@gatech.edu}% <-this % stops a space
\thanks{S. Siami-Namini is with the Department of Mathematics and Statistics, Texas Tech University, Lubbock, TX, 79409 USA e-mail: sima.siami-namini@ttu.edu}
\thanks{M. Adl Khangha and F. Mirza Soltani are with the Department of Computer Science, University of Debrecen, Debrecen, Hungary, e-mail: adl.mahdi1365@gmail.com and soltani.fahimeh1364@gmail.com}
\thanks{A. Siami Namin is with the Department of Computer Science, Texas Tech University, Lubbock, TX, 79409 USA e-mail: akbar.namin@ttu.edu}% <-this % stops a space
\thanks{The pre-print of an accepted paper for publication in the Journal of Springer Nature (SN) of Applied Sciences (March 2020).}}

\maketitle              % typeset the header of the contribution
\begin{abstract}
 Clustering is an optimization problem and an iterative process. Through clustering, observations of a given data set clustered into distinct groups. The optimization goal is to maximize the similarities of data items clustered in the same group while minimizing the similarities of data objects grouped in separate clusters. With respect to the complexity of features captured for the given data set, a simple clustering task can turn into a multi-objectives clustering process in which more than one feature is utilized to cluster data objects. Hence, the complexity of cluster analysis heavily depends on the dimensionality of the data and thus the number of features to be considered. While the dimensionality of given data is observable, the identification of possible features involved in the analysis is a challenging task. A more daunting problem is the identification of hidden features in the given data set. More specifically, the detection of such hidden features is a non-trivial task that needs more advanced algorithmic and mathematical techniques and solutions. An example of such data sets with complex structures and known and hidden features is time series data. As a special case, Time Series Clustering (TSC) inherently is a complex and multi-objective problem where the exact and complete set of features and their significance is unknown and hidden. As a result, conventional data mining-based clustering algorithms may miss these hidden features and thus not effectively perform clustering for time series data. 

Machine learning and in particular deep learning algorithms are the emerging approaches to data analysis. These techniques have transformed traditional data mining-based analysis radically into a learning-based model in which existing data sets along with their cluster labels (i.e., train set) are learned to build a supervised learning model and predict the cluster labels of unseen data (i.e., test set). In particular, deep learning techniques are capable of capturing and learning hidden features in a given data sets and thus building a more accurate prediction model for clustering and labeling problem. However, the major problem is that time series data are often unlabeled and thus supervised learning-based deep learning algorithms cannot be directly adapted to solve the clustering problems for these special and complex types of data sets.  To address this problem, this paper introduces a two-stage method for clustering time series data. First, a novel technique is introduced to utilize the characteristics (e.g., volatility) of given time series data in order to create labels and thus be able to transform the problem from unsupervised learning into supervised learning. Second, an autoencoder-based deep learning model is built to learn and model both known and hidden features of time series data along with their created labels to predict the labels of unseen time series data. The paper reports a case study in which financial and stock time series data of selected 70 stock indices are clustered into distinct groups using the introduced two-stage procedure. The results show that the proposed procedure is capable of achieving 87.5\% accuracy in clustering and predicting the labels for unseen time series data. %The results of clustering performed by the proposed autoencoder-based clustering shows that deep learning-based clustering outperforms conventional KMeans clustering in grouping stock indices.

%\keywords{KMeans Clustering  \and Financial Data Analysis \and Time-Series Clustering \and Deep Learning \and Encoder-Decoder \and Unsupervised Learning \and Supervised Learning.}

\end{abstract}

\section{Introduction}
\label{sec:intro}

\IEEEPARstart{A}{n} important step prior to performing any detailed data analysis is to understand the nature and characteristics of a given data set. There are several statistical techniques that help in creating different level of abstractions, each representing the data set from different angles. A very basic and preliminary technique is descriptive statistics such as mean and standard deviation that are often utilized by data analysts  in order to grasp the trend and variation of observations and thus capture a big picture of the data. These types of metadata can describe and, more specifically, ``{\it featurize}'' data. Hence, in practice and theory data analysis refers to the identification, selection, and analysis of the known features of data sets. 

As a very important and prevalent data type, time series data are playing a key role in different application domains such as social and human sciences such as psychology, economics, business, and finance as well as engineering, quality control, monitoring and security. What makes time series data unique is the addition of another dimension to the data (i.e., time) and the elevation of the complexity involved in analysis. The high dimensionality and additional complexity introduced by time make analysis of such data types very challenging. Therefore, it raises the concern of whether conventional data analysis techniques are suitable in exploring all ``possible'' features, factors, and their causalities of such data type. 

A popular data analysis task is the traditional clustering problem, in which the given dataset is divided into subgroups with the goal of maximizing the similarity of the data observations grouped together; while maximizing the dissimilarity of the observations clustered in distinct groups. A simple and typical clustering algorithm (e.g., KMeans) takes as input a numerical vector representing the original data and measures the distance between data items using a simple distance metric (e.g., Euclidean). The assignment of data observations to different groups is then optimized with respect to the adjustment and optimization, which is an repetitive process. It is also possible to extend the basic clustering algorithm to address a more general multi-objectives problem where more than one feature, or characteristics, of datasets are taken into account for clustering. However, the problem of clustering of time series data is undeniably more daunting and challenging that need further analysis. There are several challenges associated with the problem of Time Series Clustering (TSC) including: 

\begin{enumerate}
\item {\it Unlabeled Data}. It is hard to find or automatically label time series data. As a result, existing clustering techniques in supervised learning are less applicable. While there are great benefits associated with unsupervised learning (e.g., no need to know the exact number of desired clusters), the absence of labels in time series data would cause overlooking more accurate clustering techniques based on supervised learning, in which the labels of time series are known and thus it is easier to predict the cluster labels of time series data. 
\item {\it High Dimensionality}. Due to the inclusion of time factor, in addition to some other features, the dimensionality of the time series data and thus the number of features is increasingly high. It is therefore of utmost importance to identify salient features that contribute significantly to the characteristics of the time series data as well as reduce the effects of nuisances that exhibit themselves as false features. 
\item {\it Hidden Features}. The most critical issue is the possibility of existence of some hidden features that may not be apparent and thus are missed from the direct data analysis. Examples of such hidden factors might be exogenous and even some endogenous factors in the given data sets.  The conventional data mining and even machine learning techniques are less effective in capturing these existence but hidden features. As a result, more advanced and rigorous methods and techniques are needed to take into account these possible features when modeling the clustering solutions. 
\end{enumerate}

To address the aforementioned challenges, this paper introduces a two-stage methodology for time series clustering. The first stage of the introduced methodology targets the problem of ``unlabeled data'' in time series. The goal of this stage is to transform an unsupervised learning problem into a supervised learning and then be able to perform clustering and prediction of labels using supervised learning techniques. The basic idea is to derive the prospective cluster labels through utilization of characteristics and features of the given time series. Once the characteristics and features of time series data are ``vectorized'' (i.e., numerically calculated and represented), a conventional K-means clustering can be used to cluster the feature vector data. The generated clusters and the label for each cluster can then be utilized to label the original time series data and thus the problem can be transformed to supervised learning. 

The second stage targets the problems of ``high dimensionality'' and dealing with ``hidden features'' of time series data. The proposed approach is to utilize deep learning to capture and take into account the effects of hidden layers. Deep learning is capable of optimizing a prediction model by iteratively learning new features through various internal neural networks and the neurons incorporated in each layer. To address both problems simultaneously, an autoencoder-based deep learning algorithm is utilized, in which the autoencoder not only take into accounts the hidden features but also preserves the features that are salient for computation and prediction. Through changing the architecture of neural networks including the shape of the input data, the number of internal layers, the number of neurons on each layer, and the activation and optimization functions it is possible to enhance the accuracy of the prediction and more specifically supervised learning-based clustering problems through deep learning. The key contributions of this paper are: 
\begin{itemize}
\renewcommand\labelitemi{--}
\item Introduce a two-stage methodology to address time series clustering. In the first stage, we introduce a methodology to create cluster labels and thus enable transforming unsupervised learning to supervised learning for time series data. In the second stage, an autoencoder-based deep learning algorithm is being built to model clustering time series data is presented. 
\item Demonstrate the performance of the proposed two-stage methodology though a case study performed on clustering time series data of 70 stock indices. The results show achieving an accuracy of 87.5\% in correctly predicting the cluster labels of time series data. 
\end{itemize}

This article is organized as follows: Section \ref{sec:literature} reviews the state-of-the-art of this time series clustering. Section \ref{sec:FeatureVectors} highlights the key characteristics of time series  and in particular financial time series data. A brief overview of artificial neural networks is represented in Section \ref{sec:ANN}. The general picture of the two-stage model is presented in Section \ref{sec:synergic}, and the encoder-decoder deep learning model is decribed in Section \ref{sec:model}. The autoencoder-based model is evaluated and the results are reported in Section \ref{sec:casestudy}. Section \ref{sec:conclusions} concludes the paper and highlights future research directions.

\section{Literature Review}
\label{sec:literature}

%List of the papers that will be reviewed in this research are as follows:~\cite{aghabozorgi2015time}, ~\cite{liao2005clustering}, ~\cite{kakizawa1998discrimination},~\cite{maulik2002performance},~\cite{kovsmelj1990cross},~\cite{policker2000nonstationary},~\cite{shaw1992using},~\cite{wang2019time},~\cite{qiu2017empirical},~\cite{roelofsen2018time},~\cite{paparrizos2017fast}.

%{\color{red} Where can I find the bib of this paper https://pdfs.semanticscholar.org/0530/80a9e72656b90fdec1e438b13491544a06ed.pdf}
%

%

%Time series data is a series of data points in time order which is essentially considered as dynamic data because their feature values are changed as a function of time. 
Time series data is a series of data points in time order that their feature values are changed as a function of time. Hence, time series data is essentially considered as dynamic data.
Time-series clustering is a special type of clustering which handles grouping time-series data. In the last few decades, time-series clustering has received significant attentions~\cite{chen2007spade,ding2008querying,stefan2013move,wang2013experimental}. Time series clustering has been shown effective in extracting useful information from time-series data in various application domains.
%https://link.springer.com/chapter/10.1007/978-3-642-04921-7_41
%Look at this:http://www.cs.columbia.edu/~gravano/Papers/2015/sigmod2015.pdf
In general, time-series clustering are classified into three categories~\cite{aghabozorgi2015time} as follows:
\begin{itemize}
    \item {\it Whole time-series Clustering.} This type of clustering is used to cluster a set of individual time-series with respect to their similarity such that similar time-series are grouped into the same cluster. The notion of clustering in this approach is similar to that of conventional clustering of discrete objects.
    \item {\it Subsequence clustering.} This type of clustering is only performed on a single time-series, where the single time-series is divided into multiple segments (i.e., subsequences) using sliding window approach.
    In other words, segments or subsequences are extracted from a single time-series using a sliding window, and then clustering is performed on the extracted segments. In this approach, all the points of time-series data is assigned to different clusters.
    %sometimes in the form of streaming time-series.
    %Individual subsequences are extracted
    %using a sliding window, and then clustering is done on extracted time-series subsequences. Note that all the points of time-series data must be assigned to clusters.
    %
    \item {\it Time Point Clustering.} This category of clustering is also performed on a single time-series which is similar to the subsequence clustering model. However, in this approach it is not required to assign all points to clusters (i.e., some of them are considered as noise). The goal of this clustering is to cluster time-point instead of the whole time-series data where clustering is performed based on the combination of similarity of the data points and their temporal proximity of time.
\end{itemize}

Keogh and Lin ~\cite{keogh2005clustering} argue that subsequnce clustering is meaningless. Therefore, this section reviews only the whole time-series clustering. Various whole time-series clustering techniques have been developed in the last few decades. Most of them critically depend on the choice of distance (i.e., similarity) measure among time-series.

%Look at this:http://www.cs.columbia.edu/~gravano/Papers/2015/sigmod2015.pdf

In general, there are three different approaches to cluster whole time-series: 1) shape-based, 2) feature-based, and 3) model-based. On the other hand, with respect to the length of the time-series, whole time-series clustering can be classified into two categories: 1) shape-level, and 2) structure-level. In the shape-based approach, clustering is performed based on the shape similarity, where shapes of two time-series are matched using a non-linear stretching and contacting of the time axes. Convectional clustering methods are used in the shape-based clustering. 

In the feature-based clustering methods, feature extraction approaches are used. Feature extraction refers to the methods that transform the raw time-series into the set of features. Feature extraction is used to compress large data sets using dimensionality reduction. Hence, in feature-based clustering raw time-series are transformed into the feature vector of lower dimension (i.e., for each time-series a fixed-length and an equal-length feature vector is created). Then, a conventional clustering algorithm is applied on the lower dimension feature vector. The extracted features are usually application dependent which implies that one set of features that are useful for one application might not be relevant and useful for another one. In some studies other feature selection methods are performed to further reduce the number of feature dimensions after feature extraction~\cite{liao2005clustering}. As the notion of shape cannot be precisely defined, dozens of similarity (i.e., distance) measures have been proposed~\cite{chen2007spade,ding2008querying,stefan2013move,wang2013experimental, tavakoli2019client, tavakoli2016log}. In this paper, we also employ similar techniques in order not only to cluster but also to utilize cluster labels for further analysis and more specifically for supervised learning. 
%

%https://www.sciencedirect.com/science/article/pii/S0031320305001305
Model-based clustering approaches assume a model for each cluster  %of the clusters
and attempt to fit the data into the assumed model. Then, each raw time-series data is transformed into either model parameters (one model for each time-series) or into a mixture of underlying probability distributions. One of the major problems of the model-based approaches is the scalability problems~\cite{vlachos2004indexing} where its performance deteriorated when the clusters are very similar.

Whole time-series clustering contains four major components: 1) dimensionality reduction (or time-series representation), 2) distance measurement (or similarity), 3) the clustering algorithm, and 4) prototype definition and evaluation where prototype refers to the summerization of the time-series. Depending on the application, time-series clustering uses some or all of these components. The reason of having each component is as follows: the dimensionality reduction is usually used to fit data in memory. Afterwards, a clustering algorithm is performed on the data using a similarity (distance) measure, and as a result a prototype is created which shows a summarization of the time-series. Finally, the created prototype is evaluated using different criteria.

In the rest of this review, main components of the whole time-series clustering are discussed in more details.

\subsection{Time-series Representation Methods}
%https://www.sciencedirect.com/science/article/pii/S0031320305001305
The first major component of the whole time-series clustering is time-series representation, also known as dimension reduction. Dimensionality reduction transforms the raw time-series into another space using feature reduction methods. This component is useful as it reduces memory requirements for raw time-series and make the data fit in the main memory. In addition, it speeds up the clustering because of significantly reducing the computations required for distance calculation of the raw time-series data. The new representation is transforming the time-series to another dimensionality with reduced space such that if two time-series are similar in the original space, their representations are also similar in the reduced spaced too, i.e., a monotonic transformation. Choosing an appropriate time-series representation method plays a significant role in the efficiency and accuracy of the clustering~\cite{ratanamahatana2005novel}. As mentioned in~\cite{keogh2003need}, two major characteristics of the time-series data is high dimensionality and noise. Therefor dimensionality reduction methods can significantly increase the performance. 

To date, several time-series representation methods have been proposed to improve the performance of time-series clustering~\cite{lin2007experiencing,popivanov2002similarity}.  Ding et al.\ ~\cite{ding2008querying} have provided a comprehensive study on eight different representation methods which are performed on 38 datasets.

In general, there are four types of time-series representation methods: 1) data adaptive, 2) non-data adaptive, 3) model-based, and 4) data-dictated (or clipped data) based approaches~\cite{lin2003symbolic,ratanamahatana2005novel,bagnall2006bit,shieh2008sax} which are explained as follows:
\begin{enumerate}
    \item Data adaptive representation methods aim to minimize the global reconstruction error~\cite{wang2013experimental}, and can be applied on all types of time-series. 
    \item Non-data adaptive approaches are only appropriate for those time-series that have fixed-size and equal-length segmentation. 
    \item Model-based approaches are a special kind of time-series representation methods that are used to represent a time-series in a stochastic way such as Hidden Markov Model (HMM)~\cite{minnen2007discovering}, statistical models, time-series Bitmaps~\cite{kumar2005time}, and Auto-Regressive Moving Average (ARMA)~\cite{kalpakis2001distance}.
    \item Data-dictated (Clipped data) time-series representation approaches are the less known type of representations where the feature reduction ratio is automatically defined according on raw time-series. The most famous method of this type of representation is called clipping (bit-level) representation~\cite{bagnall2006bit}.
\end{enumerate}

\subsection{Time-series Similarity/Distance Measures Methods}
Time-series clustering are highly dependent on the choice of similarity and distance metric. An appropriate choice for similarity/distance extremely relies on the time-series representation methods, the length of the time-series, the characteristic of time-series, and the objective of clustering time-series. In general, similarity/distance measure approaches are classified into two categories: 1) clustering according to objectives, and 2) clustering according to the length of time-series which respectively require different approaches.

\subsubsection{Similarity/Distance According to Objectives} 
There are three time-series clustering objectives that are used to classify distance measures: similarity in time, similarity in shape, and similarity in change. In the following, each objective is explained in more details.
\paragraph{Finding Similar Time-series in Time}
In this approach, similar time-series are discovered on each time step. Euclidean distance and correlation based distances are appropriate distance measures for this method. However, these distance measures are calculated using raw time-series which is extremely expensive. Hence, the calculation is performed on transformed time-series such as Piece-wise Aggregate Approximation (PAA), wavelets, or Fourier transformation. For a comprehensive study on finding similar time-series in time, interested readers refer to ~\cite{keogh2003need}. 
\paragraph{Finding Similar Time-series in Shape}
Similar time-series are identified according to similar shape features regardless of time points~\cite{bagnall2005clustering}. To do so, similar trends occurring at different time or similar pattern of changes in data are captured. Elastic methods, such as Dynamic Time Warping (DTW), are used to measure distance for this approach~\cite{chu2002iterative}. Note that, similarity in time is an special case of similarity in shape. 

\paragraph{Finding Similar Time-series in Change}
Also known as structural similarity, in this approach the time-series data is first modeled using modeling methods such as Hidden Markov Models or ARMA process. Then, similarity metric is measured based on global feature extracted from the obtained models. This is an appropriate approach for long time-series (i.e., high dimensionality), and may not be effective for short or modest time-series~\cite{wang2006characteristic}.
\subsubsection{Similarity/Distance According to the Length of Time-series}
Time-series clustering according to the length of time-series is classified into two categories: 1) shape level, and 2) structure level. The shape level is used to capture similarity of short-length time-series clustering; whereas, the structure level is used for long-length clustering~\cite{wang2006characteristic}.
\subsection{Time-series Cluster Prototypes} 
One of the most significant subroutines used in time-series clustering is cluster prototype or cluster representative. Cluster prototype refers to the summarization of time-series and is obtained using different methods. The quality of clustering is highly dependent on the quality of cluster prototypes. Three main methods to obtain the cluster prototype are as following:
\begin{enumerate}
    \item Using Medoid as Prototype~\cite{kaufman2009finding}. Medoid is defined as a member of cluster such that its dissimilarity to all other members in the cluster is minimum.  The concept of medoid is similar to that of centroids (which is used in K-mean clustering) and means. However, medoids are members of cluster; whereas, centroids and means are not. Medoids are useful when centroids or means cannot be defined as graphs.
    \item Using Averaging Prototype~\cite{keogh1998enhanced}. In averaging prototype methods, mean of time-series at each point is calculated. Averaging prototype is used when the time-series have equal length and distance metric (e.g., Euclidean distance) is a non-elastic metric. Sometimes computing average of time-series is not trivial. For example, when the similarity between time-series is based on the shape, then finding the average shape is challenging so in this case averaging prototype is evaded. In general, if the similarity of time-series is based on elastic approaches (such as Dynamic Time Warping(DTW) or Longest Common Sub-Sequence (LCSS)), averaging prototype is not trivial and is evaded~\cite{gupta1996nonlinear}. 
    \item Using Local Search Prototype~\cite{hautamaki2008time}. In local search prototype, the medoid of cluster is computed, then warping paths techniques~\cite{hautamaki2008time} are used to calculate averaging prototype. Finally, for the obtained averaged prototype new warping paths are calculated. 
\end{enumerate}

\subsection{Time-series Clustering Algorithms} 
Time-series clustering are classified into six categories: 1) Hierarchical, 2) Partitioning-based clustering, 3) Density-based clustering, 4) Grid-based clustering, 5) Model-based clustering, and 6) Multi-step clustering. In the following, each clustering method is defined in more details. 
\subsubsection{Hierarchical Time-series Clustering.} 
In this approach, a hierarchy of clusters is generated using either agglomerative (or bottom-up) or divisive (or top-down) approaches. In agglomerative methods, each item is considered as a cluster then appropriate clusters are merged together; whereas, in divisive approach all the items are included in one cluster, then the cluster is split into multiple clusters. Once the hierarchy is generated, it cannot adjust with any further changes. Therefor, the quality of hierarchical clustering is weak and other clustering approaches are leveraged to remedy this issue.
\subsubsection{Partitioning Time-series Clustering.} 
In this approach, $k$ groups of clusters are generated. One of the most common algorithms of partitioning clustering is called $k$-mean clustering~\cite{macqueen1967some}, where $k$ clusters are generated and the mean value of all the elements inside a cluster is considered as a cluster prototype.
\subsubsection{Density-based Time-series Clustering.} 
In this approach, a cluster is defined as a subspace of dense objects. One of the most common algorithms of density-based clustering is called DBSCAN~\cite{ester1996density}, where a cluster is extended if its neighbors are dense. 
\subsubsection{Grid-based Time-series Clustering.} 
In the grid-based clustering, the space is divided into a finite number of cells which are called grids, then clustering is done on the grids. STING~\cite{wang1997sting} and Wave Cluster~\cite{sheikholeslami1998wavecluster} are two common grid-based clustering algorithms. 
\subsubsection{Model-based Time-series Clustering.} 
In this approach, a model is used for each cluster, then the best fit of data for the model is discovered. In model-based clustering approaches, either statistical approaches or neural network methods can be used. One example is Self-Organizing Maps (SOM) which is a model-based clustering approach based on neural networks~\cite{fu2001pattern}.  
\subsubsection{Multi-step time-series clustering} 
Multi-step time-series clustering refers to a combination of methods (also called a hybrid method), which is used to improve the quality of cluster representation~\cite{aghabozorgi2014hybrid,lai2010novel}.

\section{Feature Vectors of Time Series as Data Labels}
\label{sec:FeatureVectors}

This section first reviews the general characteristics of time series that can be used as features, and the explores financial time series and their unique characteristics along with a short description of the most representative of financial time series data: volatility and return.

\subsection{Common General Components of Time Series Data}
\label{sec:TSComponent}

General time series data are often analyzed with respect to some features and components. This section briefly presents some known features of time series: 

 \begin{itemize}
 \renewcommand\labelitemi{--}
\item {\it Seasonality}. Seasonality is a periodical pattern observed for a time series. It is the effects of seasons such as months or fiscal year on the volatility and the volume traded within a period of time. For instance, it is expected that the price of crude oil usually is elevated in the beginning of cold seasons.  
\item {\it Cycle}. Cycle is a dynamic pattern observed over a period of time (e.g., year). For instance, it is expected to observe some cyclic behavior during harvesting time (e.g., cotton harvesting time). 
\item {\it Trend}. It is a long-term movement in a given time series without considering time or some other external influential factors. For instance, it is expected that the number of individuals who purchase new Apple product increase. However, this trend will be slowly disappearing over time if another new and better product introduced into the market. 
\item {\it Irregular Features}. These types of components are unpredictable. These features are often calculated or retrieved after trend-cycle and seasonal components are removed from the time series. The remaining parts are unpredictable, since it only represents non-cyclic and the characteristics that are unique to the underlying time series.
 \end{itemize}

These features are the major tools for analyzing general time series data. More specific time series data such as those related to financial markets have their own unique features which are discussed in following section.

\subsection{Common Features of Financial Time Series data}
\label{sec:FTSComponent}

Financial time series data can be characterized through certain features and patterns \cite{Sewell}:
 \begin{itemize}
 \renewcommand\labelitemi{--}
\item {\it Dependence}. There exist a positive autocorrelation in stock return indices, but this autocorrelation is largely insignificant.  
\item {\it Distribution}. Annual returns follow a normal distribution. Security returns are non-stationary and also follow a normal distribution with fat tails. 
\item {\it Heterogeneity}. The distributions of financial returns are non-stationary. Moreover, the standard deviation of returns is not constant over time. 
\item {\it Non-linearity}. Time series models are mostly non-linear in mean and variance. 
\item {\it Scaling}. Unlike physical objects, there are no constants or absolute sizes in economics. As a result, there is no characteristic scale in economics and finance and thus financial markets demonstrate non-trivial scaling properties. 
\item {\it Volatility}. It is the standard deviation of the change in the values of a financial time series data and often used to demonstrate the risks associated with stock indices. 
\item {\it Volume}. It refers to the level of trading of a stock index over a given time period in the market. This feature may have some correlations with calendar and seasonal effects.
\item {\it Calendar Effects}. Seasonal or calendar effects are periodical anomalies or patterns that are observed in returns. There are several different types and flavors of calendar effects such as the weekend effect, the January effect, the holiday effect, and the Monday effect. 
\item {\it Long Memory}. There is a chance that stock market returns and volatility exhibit long memory properties meaning that the observed returns are dependent over time. The chance highly depends on the type of the market. 
\item {\it Chaos}. This feature exists when a dynamic system exhibits a sensitivity to initial conditions and thus reacts to unpredictable long-term behavior. There exists very small evidence of low-dimensional chaos in financial markets.
 \end{itemize}

Classical data mining techniques and even machine learning algorithms might not be able to capture all these features and thus generated model might not be accurate. As discussed and presented in this paper, deep learning approaches are better well-positioned to formulate these features through layers of learning.  

\subsection{Financial Time Series Feature Vector: $<$Volatility, Return$>$}
\label{sec:tradeoff}

In finance, volatility, also known as swings, refers to the degree of variation of a trading price series such as S\&P 500 index over time, which is calculated by the standard deviation of logarithmic returns. More specifically, volatility shows the frequency and severity in which the market price of an investment fluctuates. The stock volatility shows uncertainty of the future of the economic and financial series. The expectation of the future of economic and financial behaviors highly contributes in changing the stock volatility.

For calculating volatility, we first need to provide returns. The return of a stock in a given time period can be define as the natural logarithm of the closing price (or other series such as opening or adjusting price) at the end of the period divided by the closing price of the stock at the end of the previous period. The general equation for calculating return is as follows: 

\begin{equation}
r_{t} = ln(\frac{C_{t}}{C_{t-1}})
\label{eqn:return}
\end{equation}

Where:
\begin{itemize}
\renewcommand\labelitemi{--}
\item $r_t$ is the return of a given stock over the period, 
\item $ln$ is the natural log function, 
\item $C_t$ is the closing price at the end of the period, and
\item $C_{t-1}$ is the closing price at the end of the last period. 
\end{itemize}
For calculating the volatility, we need to calculate the standard deviation of the returns. Standard deviation is the square root of variance, which is the average squared deviation from the mean as follows:

\begin{equation}
\sigma = \sqrt{\frac{1}{T-1}}\sum_{t=1}^{T}(r_{t} - \mu)^2
\label{eqn:volatility}
\end{equation}

Where:
\begin{itemize}
\renewcommand\labelitemi{--}
\item $r_t$ is the return of a given stock over the period, 
\item $\mu$ is the average of the returns, and 
\item $\sigma$ is the square root of variance. 
\end{itemize}

As an example, VIX (Chicago Board Options Exchange Market Volatility Index) is a popular measure of the implied volatility pf S\&P index options.  If there is a wide range of fluctuations in the prices over short time, it means that there is high volatility and vice versa. On the other hand, if the price moves slowly, there is low volatility \cite{Mamtha}. 

The importance of volatility and returns and the trade-off between these two stock indicators has received tremendous attentions. In practice, investors invest in the stock markets with an expectation of getting returns, which in turn involves risks or the volatility of asset returns. In fact, the trade-off between return and risk is the conceptual framework in the asset-pricing models. 

There has been a large body of literature on transmission of stock returns and volatility. Most asset-pricing models indicate a positive trade-off between expected returns and volatility. On the other hand, there are some research studies in which empirical evidence supports a negative relationship between returns and volatility \cite{Bakaert, Whitelaw, Shawky}. For example, Chung and Chuwonganant \cite{Chung} found that market volatility affects returns through stock liquidity, suggesting that liquidity providers play an important role in the market-return relationship in the United States. Sen and Bandhopadhyay \cite{Sen} evaluated a dynamic return and volatility spillover from US stock market into the Indian stock market. These conflicting results warrant further estimation by using appropriate techniques and algorithms. 

Volatility clustering is the main feature of volatility of asset prices and the volatility shocks can affect the expectation of volatility in future \cite{Sen}. Volatility clustering means the large changes of prices (variance of return) for a period.  

There is a double relationship between volatility and returns in equity markets. Long run fluctuations of volatility show risk premiums and therefore establish a positive relation to returns. On the other hand, short run volatility indicates news effects and shocks to leverage, and thus produce a negative volatility-return relation. The leverage effect explains how the volatility rises when the asset prices reduces. While long run volatility is related with a higher return, the opposite appears in the short run volatility.

\section{Artificial Neural Networks: A Brief Review}
\label{sec:ANN}

There are several different types of deep learning-based neural networks including convolutional neural networks (CNN) and recurrent neural networks (RNN). This paper provides only a general background related to general concept of ANNs and autoencoders.   

\subsection{Artificial Neural Network (ANN)}
A typical neural network consists of different layers: 1) an input layer, 2) one or more hidden layers, and 3) an output layer. The nodes or neuron on each layer usually represent the number of features and thus the dimensionality of the datasets. The neurons are mapped through links called ``synapses'' to the nodes created in the hidden layers and then to the output layer. The synapses links are associated with some weights that represent the significance of the value hold by every node. The weights help in decision making in order to decide which feature should be considered and thus should pass through the next layers. The weights also demonstrate the strength of the features to the hidden layer. A neural network is capable of adjusting the weight for each synopsis, a process which is usually called learning through optimization.

The nodes in the internal layers utilize activation functions such as {\it sigmoid} or tangent hyperbolic ($tanh$) on the weighted sum of inputs and then transform or map the inputs to the outputs that hold the predicted values. Once the weights are adjusted, the output layer creates a vector of probabilities for different outputs and chooses the one with minimum error rate. In the case of multi-labels clustering problem, i.e., clustering with more than two outcomes, a $SoftMax$ function can be utilized in which it minimizes the differences between the expected and predicted values.

The learning process is an iterative task by which the assignments and weights are repeatedly adjusted to with the goal of minimizing the errors obtained through the network training. To find the most optimal values for errors, the errors are ``back propagated'' into the network from the output layer towards the hidden layers and as a result the weights are adjusted. The procedure is repeated several times with the same observations and the weights are re-adjusted until there is an improvement in the predicted values and subsequently in the cost. When the cost function is minimized, the model is trained.

\subsection{Encoder-Decoder}

Autoencoders are a type of neural networks that transforms input data into their output. Autoencoder uses two parts in this transformation \cite{Hubens}: 
\begin{enumerate}
\item {\it Encoder} by which it transforms its high dimensional inputs into a smaller set of dimensions while keeping the most important features, and 
\item {\it Decoder} by which the reduced set of features is used to reconstruct the initial input data. 
\end{enumerate}

The output of the encoder, referred to as ``{\it latent-space representation}'', is a compressed form of the input data in which the most influential and important features are kept. The output of the encoder is then utilized to reconstruct the initial input data given to the autoencoder. 

From mathematical point of view, an autoencoder network is a composition of functions $(f g) (x)$. More specifically, an encoder is a function $f$ that takes $x$ as input and maps $x$ into $h$, or the latent-space representation (i.e., $h = f(x)$).  On the other hand, a decoder is a function $g$ that takes the output of the encoder (i.e., $h$) and produces $r$ (i.e., $r=g(h)$). The objective is to make $r$ as close as possible to $x$. 

The key objective of autoencoders is not just to copy the input into the output. In fact, through the training of an autoencoder and transformation of the input into the output, it is aimed that the produced latent-space representation (i.e., $h$) holds only unique and important properties and features of the dataset that can be used for further analysis. In order to extract the only important features of the given dataset in the form of latent-space representation, a set of constraints can be defined on function that generates $h$ so that the resulting compressed form of the dataset has smaller dimensions than initial dataset $x$. As a result, the quality of detecting most salient features of the dataset $x$ heavily depends on the constraints defined on $h$. There are different variations of autoencoders \cite{Hubens}:

\begin{enumerate}
\item Basic autoencoder in which there are three layers: a) an input layer of size $\vert x \vert$, b) a hidden layer of size $\vert h \vert$ (i.e., $\vert h \vert < \vert x \vert$), and b) an output layer of size $\vert r \vert$ (i.e., $\vert r \vert = \vert x \vert$) in which size refers to the number of nodes incorporated and designed in the underlying layer. 

\item Multilayer autoencoder, in which the number of hidden layers is increased to more than one. This type of autoencoders is useful when additional internal hidden layers are required to extract the hidden features and train the model. 

\item Convolutional autoencoder, in which the input data is filtered for the goal of extracting only some parts of it. These types of autoencoders are particularly very effective in image processing applications and conversions from 3-D into smaller dimensions of filtered images.  

\item Regularized autoencoder, in which the extraction and training stages are performed in accordance with some other factors such as loss functions than solely based on defining hidden layers. 

\end{enumerate}

\section{A Synergic Method For Time Series Clustering} 
\label{sec:synergic}

Figure \ref{fig:flow} and \ref{fig:flow2} depict the proposed two-stage methodology for time series clustering. The methodology first enables supervised learning by generating cluster labels for the given time series data, and then it uses the generate labels for clustering of time series data. The steps of the two-stage synergic methodology is as follows:

\begin{figure*}[!ht]
    \centering
    \begin{minipage}{0.85\textwidth}
        \centering
        \includegraphics[width=0.9\linewidth]{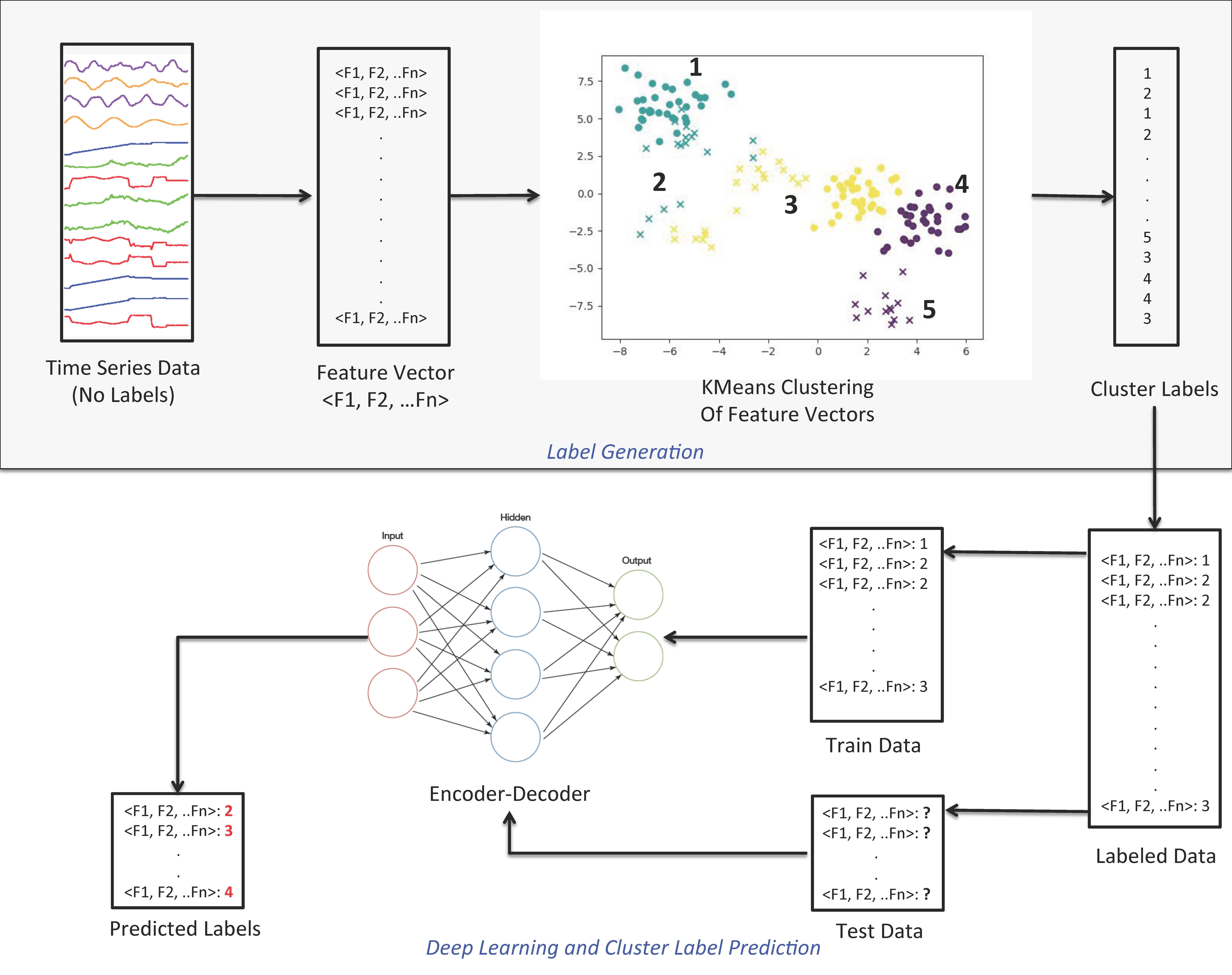}
        \caption{A synergic methodology for time series clustering.}
        \label{fig:flow}
    \end{minipage}
    \begin{minipage}{0.85\textwidth}
        \centering
        \includegraphics[width=0.9\linewidth]{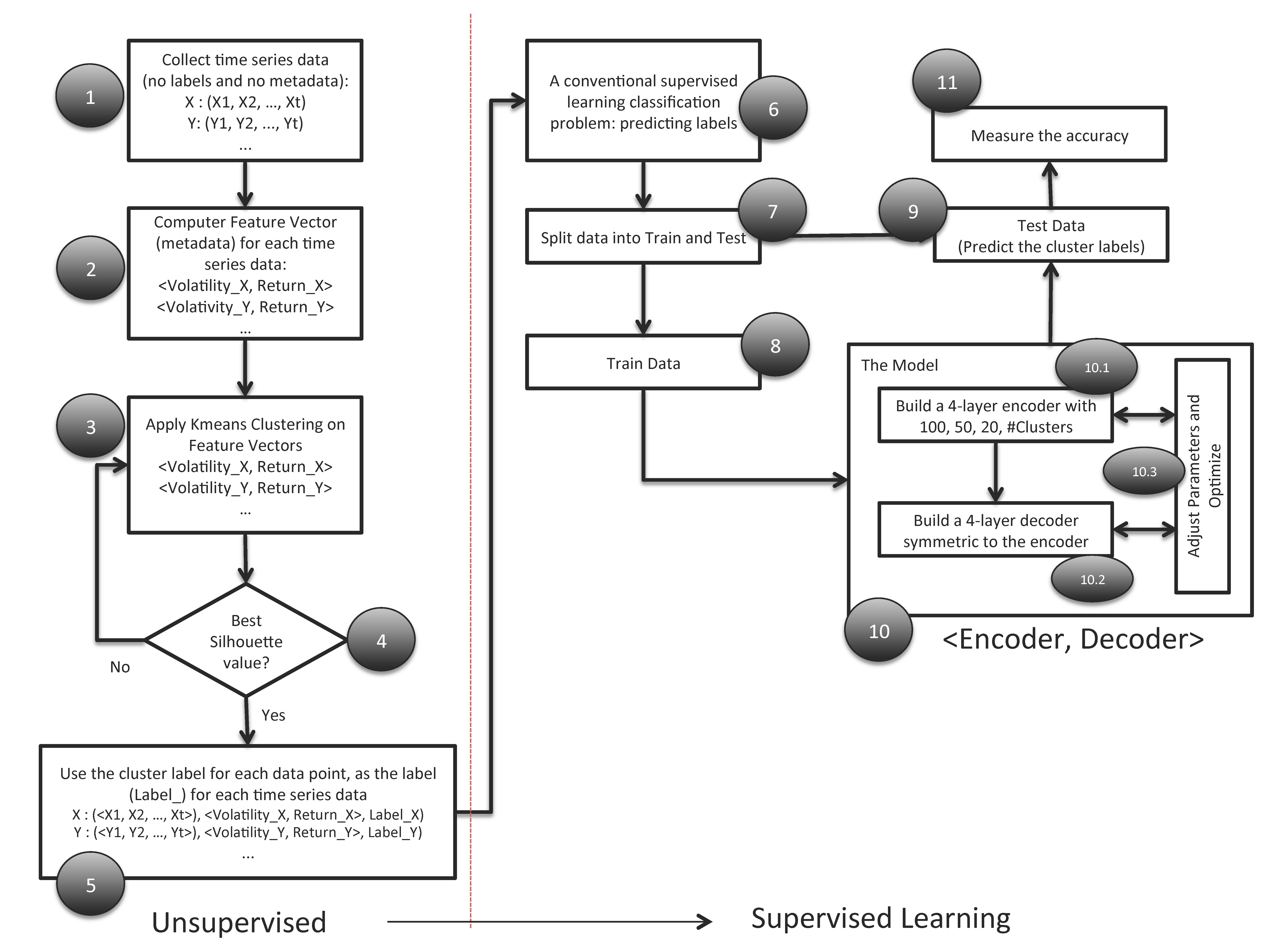}
        \caption{The flowchart of the introduced timer series clustering.}
        \label{fig:flow2}
    \end{minipage}
\end{figure*}

\paragraph{Stage I: Label Generation}
\begin{enumerate}
    \item Capture the characteristics and descriptive metadata, as features,  and build the feature vectors $<f_{1}, f_{2}, ..., f_{n}>$ for each time series data.
    \item Apply conventional KMeans clustering on feature vectors and identify cluster groups.
    \item Utilize the cluster groups and their identifications (i.e., tags) as labels for each time series data. 
    \item Provide the time series data, their feature vectors, and their generated labels to Stage II and thus transform an unsupervised learning to a supervised learning problem. 
\end{enumerate}

\paragraph{Stage II: Autoencoder-based Clustering}
\begin{enumerate}
    \item Build an autoencoder-based deep neural network with some hidden layers and neurons, i.e., nodes, in which:
    \begin{itemize}
    \renewcommand\labelitemi{--}
    \item The number of nodes on the inner most layer represents the number of clusters, 
    \item The number of nodes on the input layer represents the feature vector and its size,
    \item The number of nodes on the output layer represents a probabilistic value showing the clustering label for each data set.
    \end{itemize}
    \item Split the constructed ``labeled'' time series data into test and train datasets.
    \item Train the autoencoder-based neural network with the train dataset.
    \item Cluster and predict the labels of the test dataset using the trained neural network.
\end{enumerate}

In the following sections, we provide an in-depth description of the synergic methodology proposed for clustering time series data. First, we focus on the architecture of the designed autoencoder and then provide in-depth discussion of the algorithms developed.

\section{Encoder-Decoder for Learning Features of Time Series Data}
\label{sec:model}

\begin{figure}[!t]
  \includegraphics[width=\linewidth]{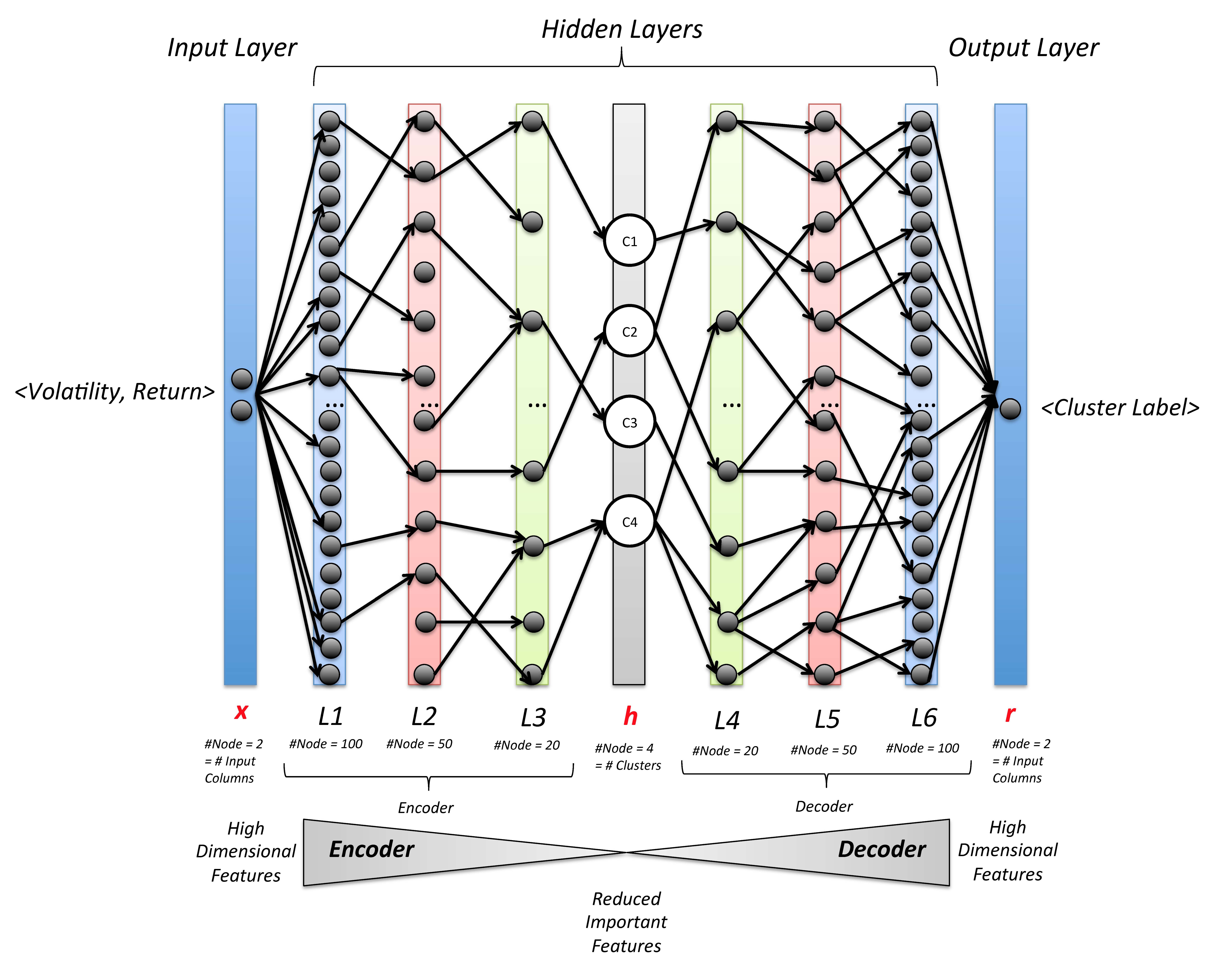}
  \caption{The designed encoder-decoder architecture.}
  \label{fig:arch}
  %\vspace{-0.15in}
\end{figure}

Figure \ref{fig:arch} demonstrates the architecture of the encoder-decoder neural network developed for feature learning of time series data. 

The network consists of two layers representing input $x$ and output $r$, respectively. The input layer is designed with two neurons (i.e., nodes) where the neurons in the input layer represent the volatility and return for each stock index (i.e., $<Volatility, Return>$, the selected features for our financial case study); Whereas, the output layer is designed with one neuron, by which a probabilistic value will be calculated to represent the cluster label.   

On the encoding part, there are three internal layers L1, L2, and L3, each with the number of devised neurons of 100, 50, and 20, respectively. Since the ultimate goal of an encoder is to reduce the dimensions of a given input, the number of nodes incorporated in these internal layers is in descending order implying the reduction of the features and preserving only those which stand out and are salient. Please note that the explicit features given to the authoencoder are in the form of $<Volatility, Return>$. However, the purpose is to detect and take into account hidden features that might exist even within volatility and return when modeling the deep learning-based clustering. 

Since an autoencoder is a symmetric neural network, the number of layers and nodes on each layer of the decoder should be symmetric with the number of layers and neurons in the encoder side. As a result, there are three layers on the decoder side (i.e., L4, L5, and L6) with the number of nodes of 20, 50, and 100, respectively, in which an ascending order of the number of neurons is apparent. 
The decoder side explicitly reconstructs the original inputs using the reduced features with exact shape for both input and output. The layer $h$ is exactly where the number of prospective clusters for clustering data is taken into consideration. In our financial case study, the optimal number of clusters is four (See Section \ref{sec:casestudy}) and thus the number of nodes on this layer (i.e., $h$) is also considered four. 

As it is apparent from Figure \ref{fig:arch}, through the number of nodes and layers defined for the encoder part, the most salient features are captured and through the decoder side, which is symmetric to the encoder side, the exact shape of the input data is reconstructed. The adjustment of weights for the internal layers and their nodes are decided and optimized in a repetitive manner where the loss function on the output (i.e., reconstructed input) is used as a means to measure the accuracy of the clustering.

The activation function incorporated on the layer $h$ is in the form of a ``{\it sigmoid}'' function. A sigmoid function is used to predict the probability values, since its values range between (0 to 1).

Once the model is trained on training set, the model (i.e., the output layer) produces a ``{\it floating}'' value in the range of $[1 - C, C - 1]$ where $C$ is the number of desired clusters (i.e., four in our case study). A simple application of rounding (i.e., {\tt np.rint} function in Python) and absolute (i.e., {\tt np.absolute} function in Python) functions to the output generated by the model will produce a ``positive integer'' value between $[0, C - 1]$ that represent the class label of the underlying stock index data for which the output has been generated.

\section{The Algorithms}
\label{sec:algorithms}

The introduced autoencoder-based deep learning methodology for time series clustering is represented through two algorithms: 1) Transforming unsupervised data into supervised through building feature vector and characterizing time series using descriptive metadata (i.e., volatility and return), and 2) Building an autoencoder-based deep learning to predict cluster labels of supervised stock data. In following sections, we describe each algorithm in further details.

\subsection{Algorithm \ref{prg:algorithm1}. Enabling Supervised Learning through Characterizing Time Series and Utilizing Metadata as Labels}

Algorithm \ref{prg:algorithm1} lists the step-by-step of transforming time series data (i.e., unsupervised data) into labeled data. The algorithm utilizes two descriptive concepts to characterize financial time series data: 1) volatility, and 2) return. Therefore, a vector of $<volatility, return>$ for each array of stock prices captured for each stock index and for a given period of time will be computed and constructed. Lets review Algorithm  \ref{prg:algorithm1} in further details. 

Algorithm \ref{prg:algorithm1} takes as inputs 1) a URL to scrap and enumerate stock indices, 2) the desired number of clusters for clustering stock indices, 3) number of stock indices to analyze and cluster, and 4) the start date of stock prices. The algorithm then labels each stock index with respect to the cluster the underlying stock index belongs to by utilizing characteristic of time series data (i.e., volatility and return) and thus converting unsupervised time series data to supervised data. 

In Algorithm \ref{prg:algorithm1}, first the setting variables are initialized (lines 1 - 5) followed by declaring with a few  data structures to hold captured data (lines 6 - 9). The algorithm then proceeds with scrapping the given URL and listing the stock indices (i.e., tickers) in order to perform cluster analysis (lines 10 - 11). The case study performed and reported in this paper focuses only on the first 70 stock indices. Once the list of stock tickers is prepared, the ``Adjusted Close'' price of each stock index is retrieved for a given time period. For our case study, we retrieved data for the start date of January 1, 2019 to April 15, 2019 (the day of running this experiment and capturing the data) (lines 12 - 15).

The constructed and filled data structure (i.e., $TP\_DF$: $<ticker, prices []>$) is then sorted in order to enable one to one cluster assignment of each stock index (line 16). As two of the key characteristic features of time series, the volatility and return values are computed for each time series of prices for each stock index and the computed values along with the stock indices are preserved in a data structure (i.e., $TVR\_DF$) (lines 17 - 22). Then, a projection of $<index, volatility, return>$ data saved in $TVR\_DF$ is created for the purpose of KMeans clustering and creating labels based on $<volatility, return>$ (line 23). 

A KMeans-based clustering model with respect to the number of desired clusters (i.e., $no\_clusters$) is then built (line 24) and the captured $<volatility, return>$ are then given to the clustering model in order to cluster and then create cluster labels (line 25). The centroids of clusters are optimized and the silhouette value for the clustering is computed (lines 26 - 27). The labels for each stock index is created, representing the cluster they belong to (i.e., 0, 1, 2, 3). The data are then saved in a file that will be used by Algorithm \ref{prg:algorithm2} to build an autoencoder.

\lstset{basicstyle=\scriptsize\ttfamily}
\def\listingsfont{\ttfamily} 
\begin{lstlisting}[float=*, language=Java, label=verb1,caption={Enabling supervised learning through utilization of metadata of time series as labels.}, frame=tb, label={prg:algorithm1}, frame=shadowbox, escapeinside=`']
Algorithm 1 (Transforming Unsupervised Learning to Supervised Learning): 
Description: Transforming Unsupervised Data to Labeled Data and Clustering 
Stock Labeled Data Using Optimal KMean Algorithm.
Inputs: 1) URL to scrap, 2) Number of Clusters, 
    3) Number of Stock Tickers, 4) The Start Date of Collecting Stock Prices  
Outputs: The Cluster Label of the Scrapped Stock Tickers

# Setting 
1.  numpy.random.seed(7)       # For reproducibility purpose
2.  sp500_URL = 'https://en.wikipedia.org/wiki/List_of_S%26P_500_companies'   # Web page to scrape
3.  no_clusters = 4    # Number of desired clusters obtained through experiments
4.  no_tickers  = 70           # Number of tickers to scrap and analyze
5.  start_date  = ‘01/01/2019’ # The start date to scrap tickers’ data 

# Declaring Some Data Frames to Hold Scrapped Data 
6.  tickers  = []  # A data frame to hold the tickers: <ticker>
7.  TP_DF    = []  # A data frame to hold the scrapped data: <ticker, prices[]>
8.  TVR_DF   = []  # A data frame to hold: <ticker, volatility, returns> 
9.  VR_DF    = []  # A data frame to hold: <volatility, returns> (used by clustering)

# Scrapping the Web page and tickers
10.  sp500_scrapped = read_html(sp500_URL)  # using Panda’s read_html function
11. tickers = read(sp500_scrapped)

# Retrieve “Adj Close” from yahoo regarding each ticker 
12. for each ticker in tickers[<= no_tickers] do
13.    prices = read(ticker, ‘yahoo’, start_date)[‘Adj Close’]
14.    TP_DF.append(<ticker, prices[]>);
15. end for
# Sort the records to re-construct based on “ticker” or index
16. TP_DF.sort

# Compute volatility and returns regarding each ticker
17. for each ticker in TP_DF do
18.    index      = ticker
19.    returns    = mean(prices) * 252
20.    volatility = std(prices) * sqrt(252)
21.    TVR_DF.append(<index, volatility, returns>)
22. end for 

# Build VR_DF[] using TVR_DF[]
23. VR_DF = <TVR_DF.volatility, TVR_DF.returns>

# Cluster <returns, volatility> data without ticker using KMean clustering
# Build the clustering model using KMean
24. clusters = KMean(n_clusters = no_clusters)
# Fit the model/predict the cluster labels regarding each data item <ret, vol>
25. predicts = clusters.fit_predict(VR_DF)
# Report the silhouette value using Euclidean distance and identify centroids
26. centers = clusters.cluster_center_
27. score = silhouette_score(VR_DF, predicts, metric = ‘euclidean’)

# Assign the cluster tag regarding each ticker (<index, voly, ret, cluster>)
28. for each ticker in TP_DF do
29.    TVR_DF.cluster[index == ticker] = pd.DataFrame(predicts[index == ticker])
30. end for

# Save the data to a file to be used by Algorithm 2: (<ticker, vol, ret, cluster>)
31. TVR_DF.to_csv("/.../k-means-StockData.csv")
\end{lstlisting}

\subsection{Algorithm \ref{prg:algorithm2}. Predicting Cluster Labels of Time Series Data through Autoencoder-based Deep Learning}

The second part of the methodology builds an autoencoder-based deep learning for clustering stock indices. The algorithm takes as inputs: 1) training labeled time series data, 2) testing unlabeled data, 3) number of clusters (i.e., neuron or node) to encode, 4) the shape of the input data (i.e., 2 in our case $<volatility, return>$, 5) the shape of the output data (i.e., 1 in our case, a floating value), 6) the number of iterations or epochs. The detailed of Algorithm \ref{prg:algorithm2} is given in Listing \ref{prg:algorithm2}.

\lstset{basicstyle=\scriptsize\ttfamily}
\def\listingsfont{\ttfamily} 
\begin{lstlisting}[float=*, language=Java, label=verb1,caption={Deep learning-based (Encoder-Decoder) supervised learning for predicting cluster labels of stock indices.}, frame=tb, label={prg:algorithm2}, frame=shadowbox, escapeinside=`']
Algorithm 2 (Supervised Learning through Autoencoder-based Deep Learning): 
Description: Building An Autoencoder to Predict Cluster Label of Stock Indices 
Inputs: 1) Training labeled set, 2) Testing unlabeled set. 
Outputs: An Autoencoder-based Deep Learning Model to Predict Cluster Labels 

# Setting 
1.  seed = 7
2.  numpy.random.seed(seed)    # For reproducibility purpose
3.  no_clusters = 4    # Number of desired clusters (i.e., # of Neurons or Nodes)
4.  BatchSize   = 1024 # data batch size retrieved by the learner in each iteration 
5.  InCol = 2  # The shape of the input data used regarding training <vol, ret>
6.  OuCol = 1  # The shape of the output data, A floating value
7.  TestSize = 0.33  # the percentage of the “test” data of the splitting data 
8.  noEpochs = 1000 # Number of epochs (learning round) regarding training the model

# Loading labeled data (train and test): (<ticker, volatility, returns, cluster>)
9.  TVR_DF = pd.read_csv("/…/k-means-StockData.csv")  # Created by Algorithm 1

# Splitting the data set into training and test set
10.  x = TVR_DF[<volatility, returns>]
11.  y = TVR_DF[<cluster>]
12.  X_train, X_test, y_train, y_test = 
                 train_test_split(x, y, test_size=TestSize, random_state = seed)

# Alternatively the InCol = TVR_DF.shape[1] command can be used to capture the 
# input shape, instead of using the hard coding style regarding InCol. 
# InCol = TVR_DF.shape[1]

# Build a tensor shape
13. input_dim = Input(shape = (InCol, )) 

# Build the autoencoder as shown in Figure XXX
# Build the encoder part that represents the input
14. encoded = Dense(100, activation = ‘relu’)(input_dim)
15. encoded = Dense(50, activation = ‘relu’)(encoded)
16. encoded = Dense(20, activation = ‘relu’)(encoded)
17. encoded = Dense(no_cluster, activation = ‘sigmoid’)(encoded) 

# Build the decode part that losey reconstruct the input 
18. decoded = Dense(20, activation = ‘relu’)(encoded)
19. decoded = Dense(50, activation = ‘relu’)(decoded)
20. decoded = Dense(100, activation = ‘relu’)(decoded) 
21. decoded = Dense(OuCol)(decoded) 

# Map input to its reconstruction
22. autoencoder = Model(input_dim, decoded)

# Compile the autoencoer with proper optimizer and loss function
23. autoencoder.compile(optimizer='adam', loss='mse')

# Train the autoencoder model using training data set 
24. train_history = autoencoder.fit(X_train, y_train, 
                 epochs = noEpochs, batch_size = BatchSize)   

# Predict the cluster tag of the test data set using the autonecoder model
25. predicts = autoencoder.predict(X_test)

# Report the labels of each stock indices in the test data
26. return np.absolute(np.rint(predicts))

\end{lstlisting}

The algorithm starts with initiating setting variables including: 1) the number of clusters to project (i.e., $no\_cluster$), 2) the number of batch size to retrieve and feed the autoencoder (i.e, $BatchSize$), 3) the shape of the input data (i.e., in our case is 2, which is the number of input columns entered to the model ($<volatility, return>$)), 4) the output shape (i.e., in out case is 1, an output with one column, which is a floating variable representing the cluster label), 5) the test size (i.e., 33\% for testing and 67\% for training), and 6) the number of epochs for iterative training (lines 1  - 8). The algorithm then loads previously saved data that were captured through Algorithm \ref{prg:algorithm1} and saves the data into a data structure $TVR\_DF$ (line 9).  The loaded data are then split into two data sets of train (68\%) and  (33\%) for test sets (lines 10 - 12). 

The exact building of the autoencoder starts with specifying the shape of the input data (line 13). In our case, the shape of the input data (i.e., $InCol$) is a vector with two columns $<volatility, return>$. The creation of different layers of the autoencoder starts at line 14 where the input shape is given to build the $x$ part of the autoencoder model (as specified in Figure \ref{fig:arch}). The input layers of the autoencoder are built through lines 14 -- 17 where the shape of the input (i.e., $input\_dim$) is given to the first layer (line 14) with 100 neurons, and the built first layer and its output is given to the second (line 15) with 50 nodes or neurons, and the third (line 16) with 20 neurons or nodes. The activation function for building these layers is ``{\it relu}'' which returns a value between (0 to 1). The $h$ part of the autoencoder (Figure \ref{fig:arch}) is built by line 17, where number of cluster labels is specified. The activation function here is ``{\it sigmoid}.''

The encoding part of the autoencoder and the encoding layers are built through lines 14 - 17 in which a decreasing number of neurons or nodes on each layer indicates focusing on important features of data and preserving them for further analysis by the next layer (i.e., feature reduction).

Conversely and in a similar manner, the decoder part of the autoencoder intends to reconstruct the initial input data using the encoded data (lines 18 - 21). The first layer of the decoder takes the output of the ``encoder'' with 20 neurons (line 18). The additional decoder layers are then built symmetrically with respect to the layers incorporated for the encoder part (lines 18 - 21) with similar activation function. Eventually, the $r$ part of the autoencoder (See Figure \ref{fig:arch}) is build where a floating variable is estimated to show the cluster label of the input data $<volatility, return>$.  

The built autoencoder maps the input to the decoded and reconstructed output and the model itself is built (line 22). The model is then compiled using ``{\it adam}'' optimizer and mean square error (i.e., mse) as a metric to assess the precision of the prediction (line 23). The built model is then given the training data set (line 24) with a given number of epochs and bath size and eventually the test data are provided to the model for the purpose of prediction of their cluster labels (line 25). In the end, the absolute and the round value of the floating output value is reported as the predicted cluster label (line 26). 

\section{Case Study and Evaluation} 
\label{sec:casestudy}

This section reports the results of a case study performed and evaluates the introduced two-stage synergic methodology to cluster financial time series data.

\subsection{Development Platform} 
The authors implemented the algorithms in Python 2.7.13, the anaconda version. The deep learning portion of the algorithms was developed using tensorflow and keras, the open source Python implementations of deep learning and neural networks. The experiments were executed on a Mac computer with OS X El Capital 10.11.2 operating system with 2.8 GHz Intel Core i7 and 16GB 1600 MHz DDR3. 

\subsection{Data Collection}

The authors collected the indexes and ticker symbols for 70 companies listed by S\&P 500. The ticker symbols were scarped from the URL of the Wiki page of the S\&P 500\footnote{https://en.wikipedia.org/wiki/List\_of\_S\%26P\_500\_companies}. The $read\_html$ Python library was used to automatically scrap and extract the required data from the given Web page. Once the thicker and symbol of the selected companies are identified, a Python script captured the time series data and more specifically the ``Adjusted Close'' for the selected stock symbol. Moreover, the adjusted close value data were captured for the period of January 1, 2019 to April 15, 2019 on a daily basis. 

\subsection{The Optimal Number of Clusters}
\label{subsec:optimal}

The determination of the optimal number of clusters is essential in improving the precision and accuracy of the proposed algorithm. An optimal clustering groups time series data with respect to an optimization metric and assigns the best label for each time series data that can be used in later stages of the algorithm for training and testing. There are several known methods to determine the best number of clusters that best clusters data with respect to the optimization metric. The Elbow method, Average Silhouette method, and Gap statistics method are a few methods in finding the optimal number of clusters. 

\begin{figure}[!t]
  \includegraphics[width=\linewidth]{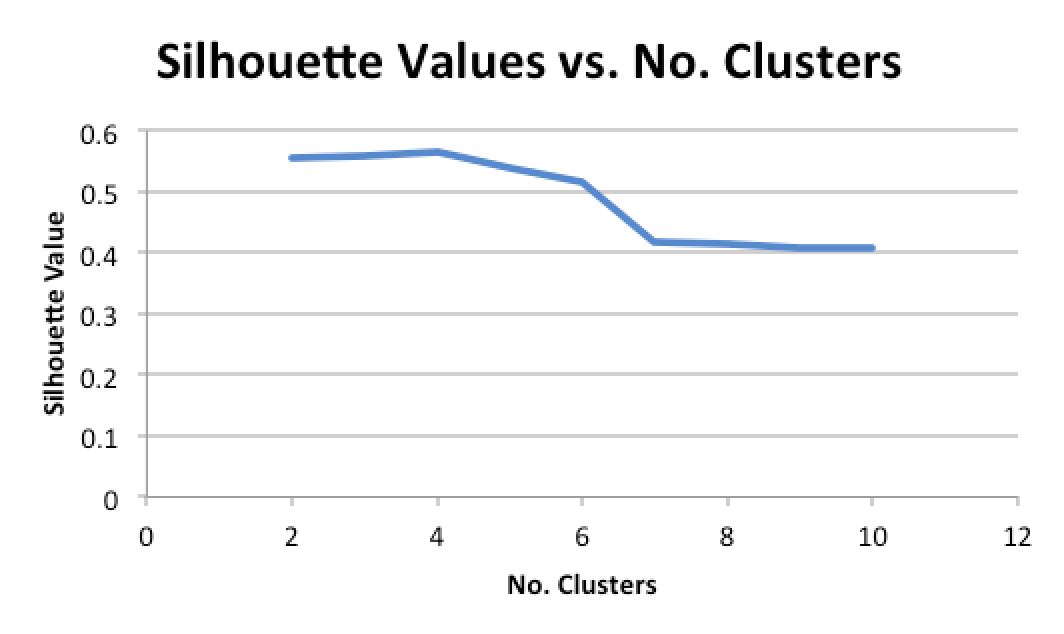}
  \caption{Optimal number of clusters.}
  \label{fig:sil}
  %\vspace{-0.15in}
\end{figure}

The authors used average silhouette method to decide about the optimal number of clusters. To do so, the conventional KMeans clustering algorithm with a desired number of clusters between 2 and 10 was applied to the feature vector data set. Figure \ref{fig:sil} illustrates the obtained Silhouette value for each clustering with different number of clusters.

As Figure \ref{fig:sil} shows the best optimal Silhouette value is produced when the number of clusters is set to 4 (i.e., Silhouette Value $= 0.564$). Therefore, the authors set the number of clusters to $4$ for the remaining part of the case study.

\subsection{Building Feature Vector: Capturing Descriptive Metadata}

Stock market data and their time series can be characterized through two concepts: 1) volatility, and 2) return. In addition to some other relevant concepts, volatility and return can be utilized to summarize the trend and certain behavior of time series. This section describes how these two characteristics are calculated and used in the clustering of time series data. 

{\it a) Annualized Stock's Volatility\footnote{https://www.fool.com/knowledge-center/how-to-calculate-annualized-volatility.aspx}}.  To calculate annualized stock's volatility, the standard deviation of price should be multiplied by the square root of $252$ assuming that there are $252$ trading days in a given year.

{\it b) Annualized Stock's Return}. The annualized stock's return is computable in a similar fashion. However, instead of standard deviation, the mean value of prices should be multiplied by the square root of $252$.

\subsection{Creating Time Series Clusters with KMeans Clustering}

Once the annualized stock's volatility and return values are computed for each stock data, the values will be given to a conventional KMeans clustering algorithm with a desired number of clusters identified before (i.e., 4). The KMeans algorithm will group stock's data with respect to volatility and return using ``Euclidean'' distance measure. The label of clusters formed by the KMeans algorithm will be used as the label for each time series data resulting in the clustering problem of unlabeled data (i.e., unsupervised learning problem) to be transformed to a clustering problem with labels and thus a supervised learning problem.  

Table \ref{tab:KMeans} lists the exact values for volatility and return for each member along with the mean and standard deviation values of these features for each cluster.

\begin{table*}[!t]
\caption{The result of KMeans clustering based on four clusters.}
\label{tab:KMeans}
\begin{center}
%\begin{tabular}{p{7.5cm}}
\begin{tabular}{|c|c|r|r|c|r|r|c|r|r|c|r|r|}
\hline
 & \multicolumn{3}{|c|}{\bf Cluster "0"} & \multicolumn{3}{|c|}{\bf Cluster "1"} & \multicolumn{3}{|c|}{\bf Cluster "3"} & \multicolumn{3}{|c|}{\bf Cluster "2"} \\
\hline
& {\bf Index} & {\bf Vol.} & {\bf Ret.} & {\bf Index} & {\bf Vol.} & {\bf Ret.} & {\bf Index} & {\bf Vol.} & {\bf Ret.} & {\bf Index} & {\bf Vol.} & {\bf Ret.} \\
\hline
 1 & ACN & 0.179 & 0.909 & MMM & 0.201 & 0.512 & AMD & 0.71 & 1.651 & ABBV & 0.254 & -0.234 \\
 2 & ADBE & 0.223 & 0.713 & ABT & 0.207 & 0.468 & ALXN & 0.298 & 1.229 & ABMD & 0.398 & -0.416 \\
 3 & AES & 0.172 & 0.909 & AAP & 0.29 & 0.516 & ALGN & 0.427 & 1.431 & ATVI & 0.46 & 0.154 \\
 4 & A & 0.2 & 0.781 & AMG & 0.296 & 0.537 & APC & 0.696 & 1.402 & ALK & 0.259 & -0.000 \\
 5 & APD & 0.162 & 0.741 & AFL & 0.109 & 0.328 & APTV & 0.291 & 1.478 & ABC & 0.283 & 0.07 \\
 6 & AKAM & 0.203 & 0.982 & ALB & 0.333 & 0.317 & ANET & 0.377 & 1.634 & AMGN & 0.209 & 0.04 \\
 7 & ARE & 0.136 & 0.969 & ALLE & 0.178 & 0.582 &  &  &  & ANTM & 0.338 & 0.035 \\
 8 & AMT & 0.123 & 0.866 & AGN & 0.31 & 0.299 &   &   &   &   &   &  \\
 9 & AMP & 0.256 & 1.05 & ADS & 0.317 & 0.612 &   &   &   &   &   &  \\
 10 & AME & 0.181 & 0.888 & LNT & 0.137 & 0.503 &   &   &   &   &   & \\ 
 11 & APH & 0.208 & 0.978 & ALL & 0.136 & 0.65 &   &   &   &   &   &  \\
 12 & ADI & 0.285 & 1.08 & GOOGL & 0.223 & 0.557 &   &   &   &   &   &  \\
 13 & ANSS & 0.204 & 1.036 & GOOG & 0.225 & 0.573 &  &  &  &   &   &  \\
 14 & AON & 0.254 & 0.766 & MO & 0.278 & 0.584 &  &  &  &   &   &  \\
 15 & AOS & 0.201 & 0.918 & AMZN & 0.284 & 0.689 &  &  &  &   &   &  \\
 16 & APA & 0.336 & 1.157 & AEE & 0.14 & 0.483 &  &  &  &   &   &  \\
 17 & AIV & 0.127 & 0.708 & AAL & 0.379 & 0.317 &  &  &  &   &   &  \\
 18 & AAPL & 0.308 & 0.894 & AEP & 0.125 & 0.553 &  &  &  &   &   &  \\
 19 & AMAT & 0.4 & 0.997 & AXP & 0.152 & 0.571 &   &   &   &   &   &  \\
 20 & ADSK & 0.287 & 1.076 & AIG & 0.276 & 0.617 &   &   &   &   &   &  \\
 21 & ADP & 0.169 & 0.851 & AWK & 0.123 & 0.6 &   &   &   &   &   &  \\
 22 & AZO & 0.202 & 0.866 & ADM & 0.178 & 0.253 &   &   &   &   &   &  \\
 23 & AVB & 0.102 & 0.708 & ARNC & 0.396 & 0.489 &   &   &   &   &   &  \\
 24 & AVY & 0.186 & 0.959 & AJG & 0.157 & 0.439 &   &   &   &   &   &  \\
 25 & BHGE & 0.278 & 0.879 & AIZ & 0.165 & 0.258 &   &   &   &   &   &  \\
 26 & BLL & 0.163 & 0.963 & ATO & 0.14 & 0.457 &   &   &   &   &   &  \\
 27 & BAC & 0.256 & 0.733 & T & 0.182 & 0.443 &   &   &   &   &   &  \\
 28 & BAX & 0.152 & 0.721 & BK & 0.183 & 0.417 &   &   &   &   &   &  \\
 29 &   &   &   & BBT & 0.213 & 0.426 &   &   &   &   &   &  \\
 \hline
\multicolumn{2}{|c|}{\bf Mean} & 0.212 & 0.896 &   & 0.218 & 0.484 &   & 0.314 & -0.050 &   & 0.466 & 1.470 \\
\multicolumn{2}{|c|}{\bf STD} & 0.069 & 0.127 &   & 0.081 & 0.119 &   & 0.089 & 0.200 &   & 0.190 & 0.157 \\
\hline
\end{tabular}
\end{center}
\end{table*}

As Table \ref{tab:KMeans} reports clusters 0, 1, 2, and 3  have 23, 30, 5, and 12 members, respectively. The mean values for the pair of $<volatility, return>$ for each cluster 0, 1, 2, and 3 are $<0.212, 0.896>$, $<0.218, 0.484>$, $<0.314, -0.050>$, and $<0.466, 1.470>$, respectively.

\begin{figure*}[t]
    \centering
    \begin{minipage}{0.45\textwidth}
        \centering
        \includegraphics[width=0.9\linewidth]{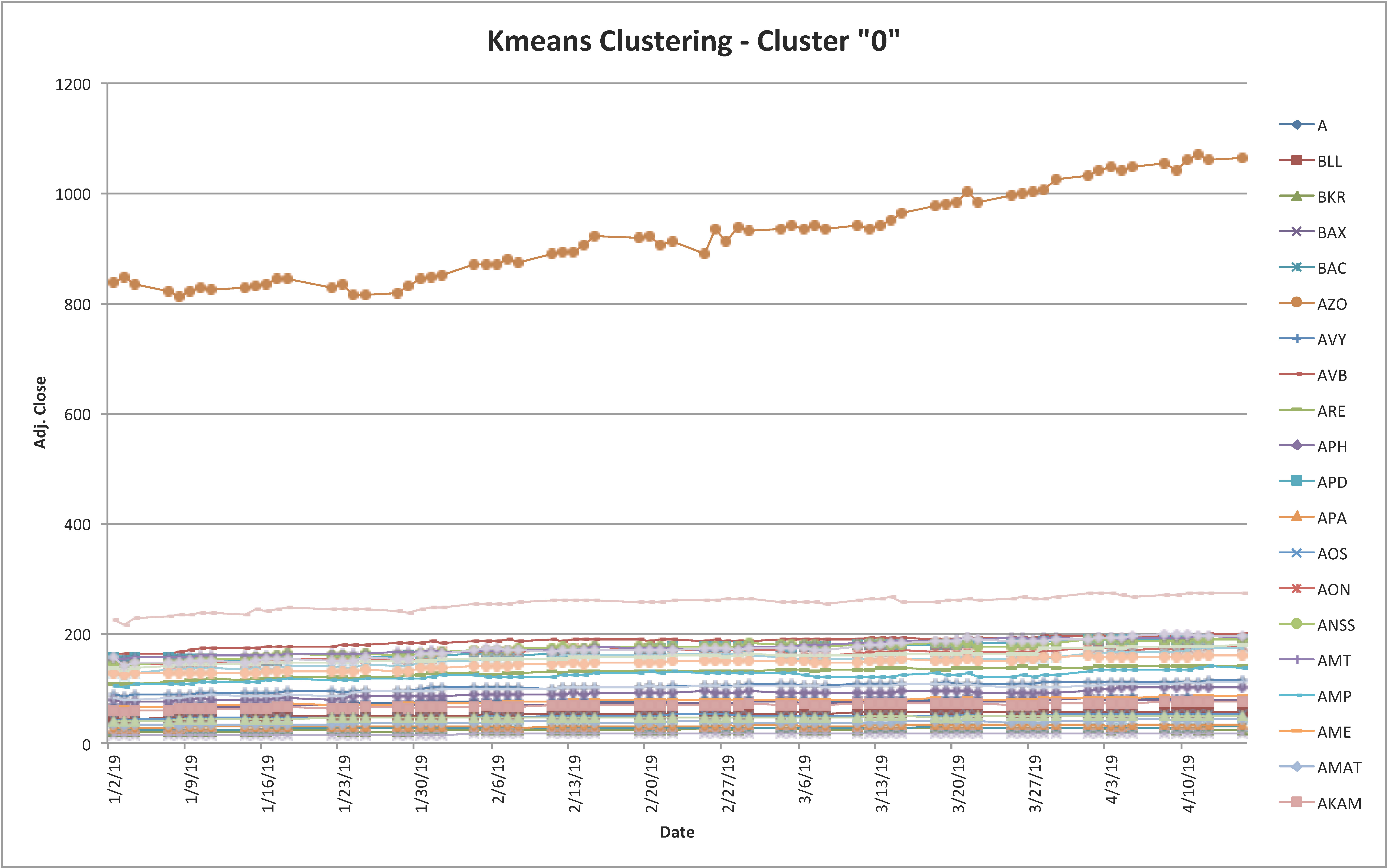}
        \caption{KMeans clustering: Cluster "0"}
        \label{fig:kmeans-0}
    \end{minipage}%
    \begin{minipage}{0.45\textwidth}
        \centering
        \includegraphics[width=0.9\linewidth]{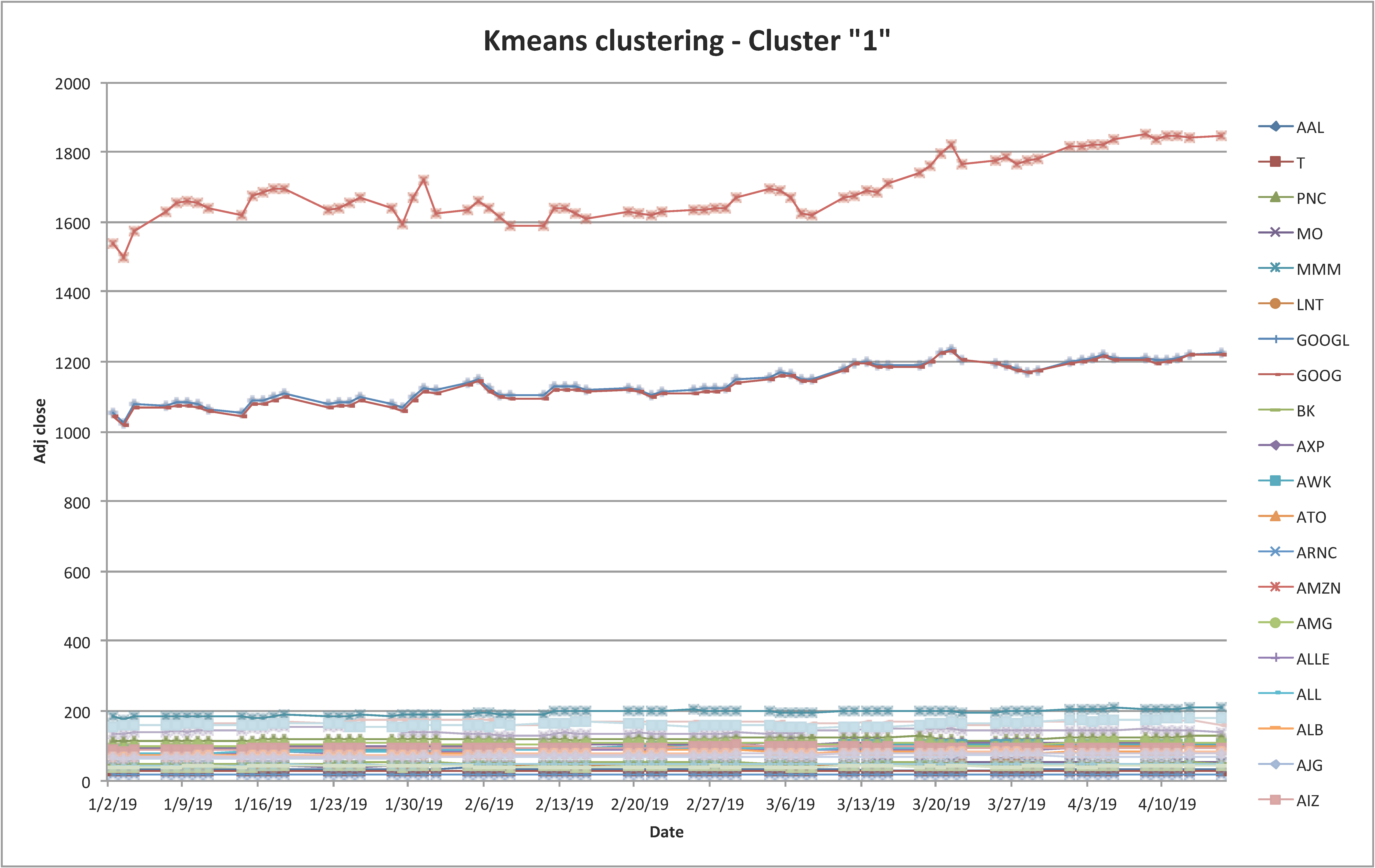}
        \caption{KMeans clustering: Cluster "1"}
        \label{fig:kmeans-1}
    \end{minipage}
        \begin{minipage}{0.45\textwidth}
        \centering
        \includegraphics[width=0.9\linewidth]{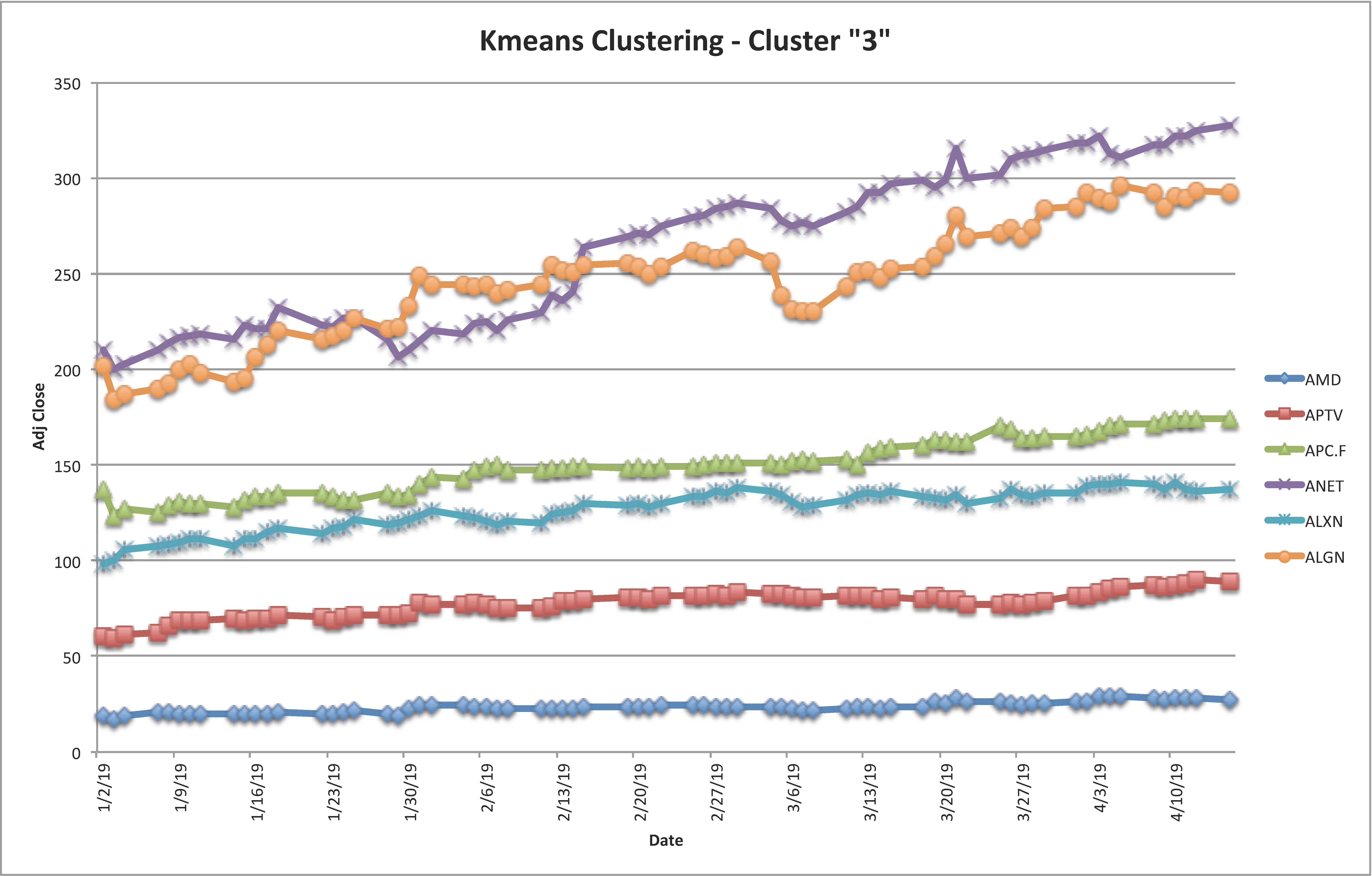}
        \caption{KMeans clustering: Cluster "3"}
        \label{fig:kmeans-2}
    \end{minipage}
        \begin{minipage}{0.45\textwidth}
        \centering
        \includegraphics[width=0.9\linewidth]{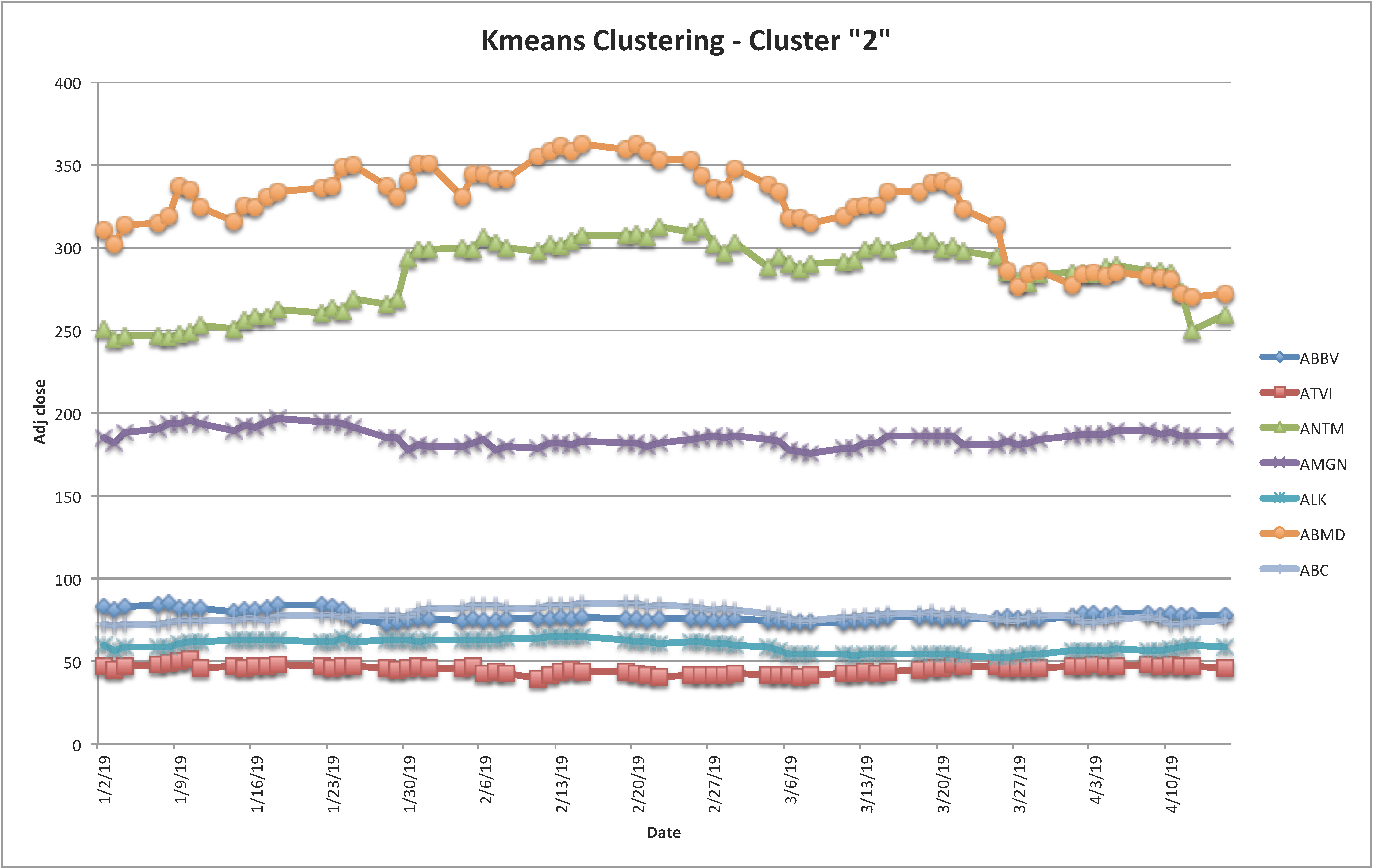}
        \caption{KMeans clustering: Cluster "2"}
        \label{fig:kmeans-3}
    \end{minipage}
\end{figure*}

To help understanding the results of the KMeans clustering, we visualize the time series data for each member  along with the range of volatility and return for each cluster. Figures \ref{fig:kmeans-0} - \ref{fig:kmeans-3} illustrate the time series data clustered together. As the Silhouette analysis indicated the optimum number of clusters to be four, the figures show the exact time series of members of each cluster for the period of January 1, 2019 and April 15, 2019\footnote{Yahoo Finance (https://finance.yahoo.com/) was used to draw the charts.}.

Let us take a look at the clusters and the stock indices grouped together. 

\begin{figure*}[t]
    \centering
    \begin{minipage}{.45\textwidth}
        \centering
        \includegraphics[width=0.9\linewidth]{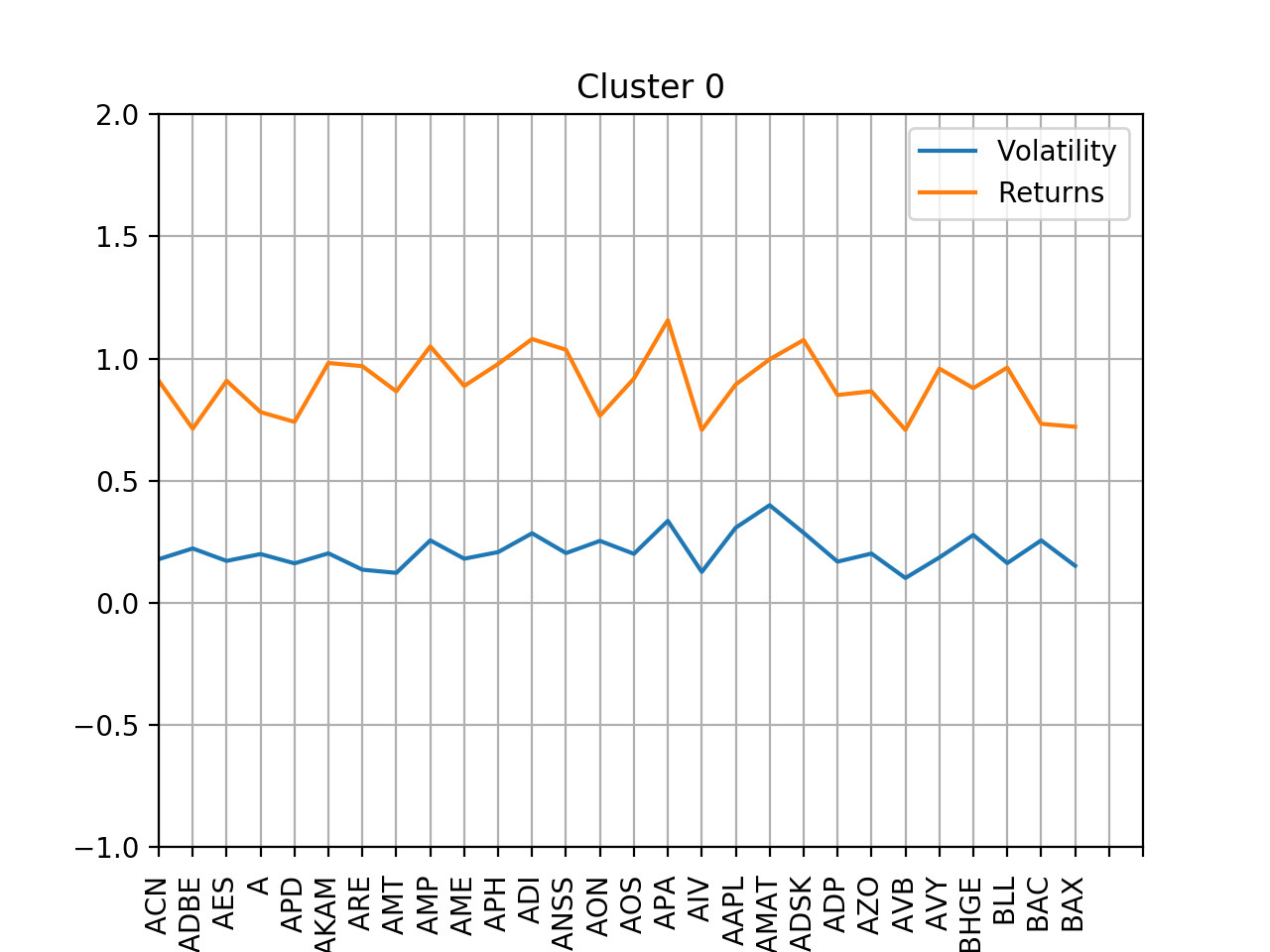}
        \caption{KMeans clustering: The range of volatility and returns for Cluster "0"}
        \label{fig:kmeans-0-VR}
    \end{minipage}%
    \begin{minipage}{0.45\textwidth}
        \centering
        \includegraphics[width=0.9\linewidth]{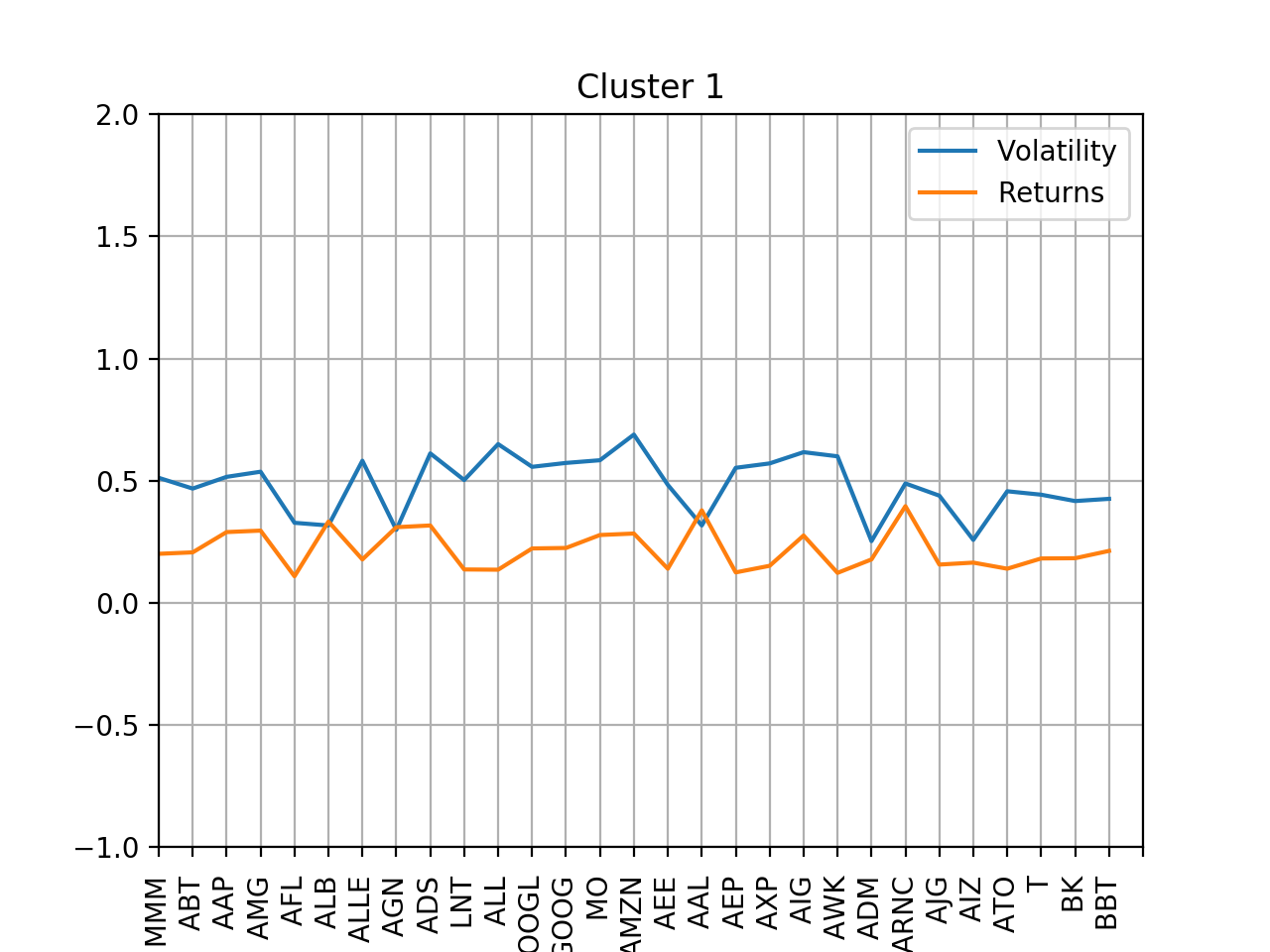}
        \caption{KMeans clustering: The range of volatility and returns for Cluster "1"}
        \label{fig:kmeans-1-VR}
    \end{minipage}
        \begin{minipage}{0.45\textwidth}
        \centering
        \includegraphics[width=0.9\linewidth]{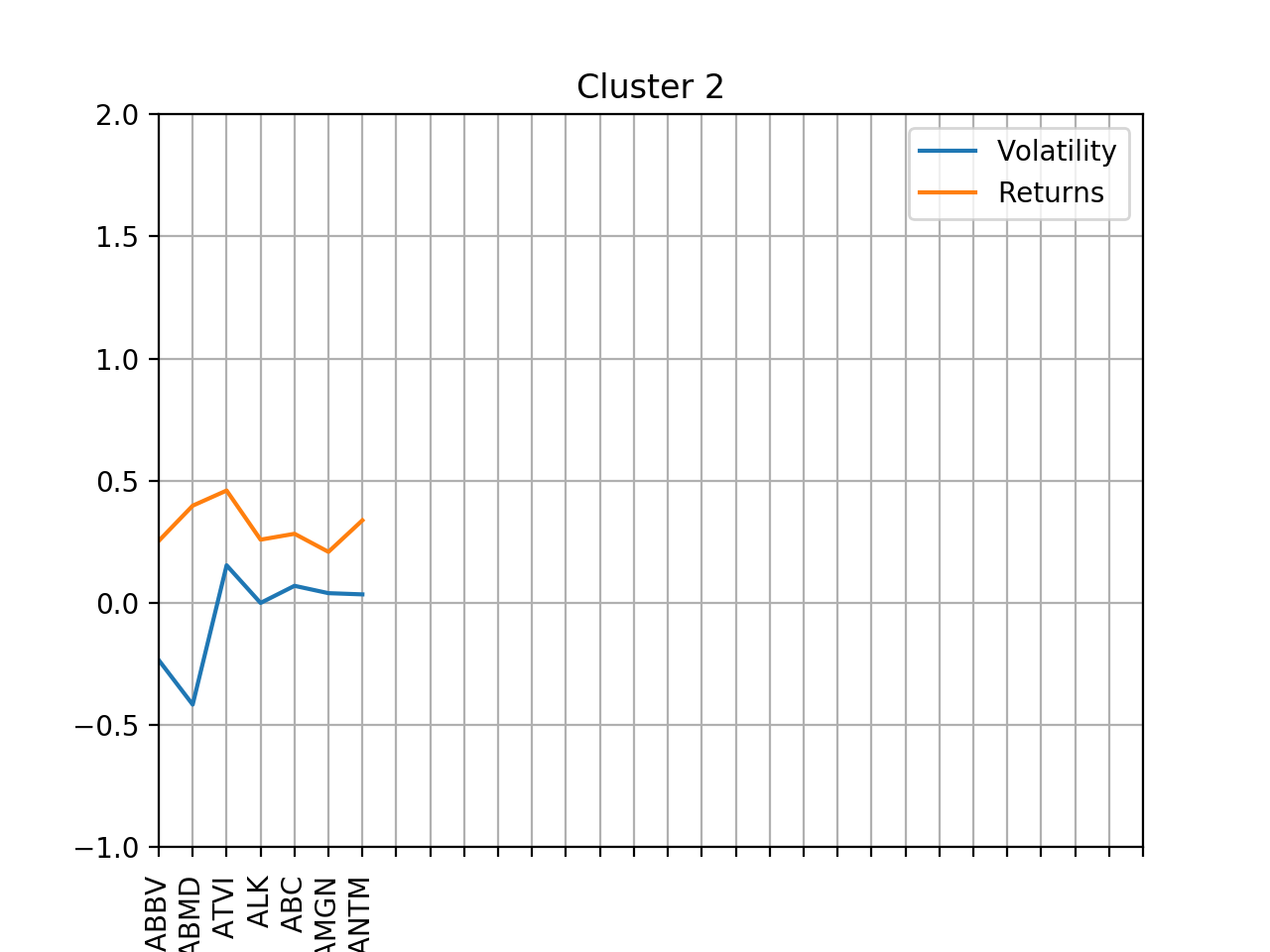}
        \caption{KMeans clustering: The range of volatility and returns for Cluster "2"}
        \label{fig:kmeans-2-VR}
    \end{minipage}
        \begin{minipage}{0.45\textwidth}
        \centering
        \includegraphics[width=0.9\linewidth]{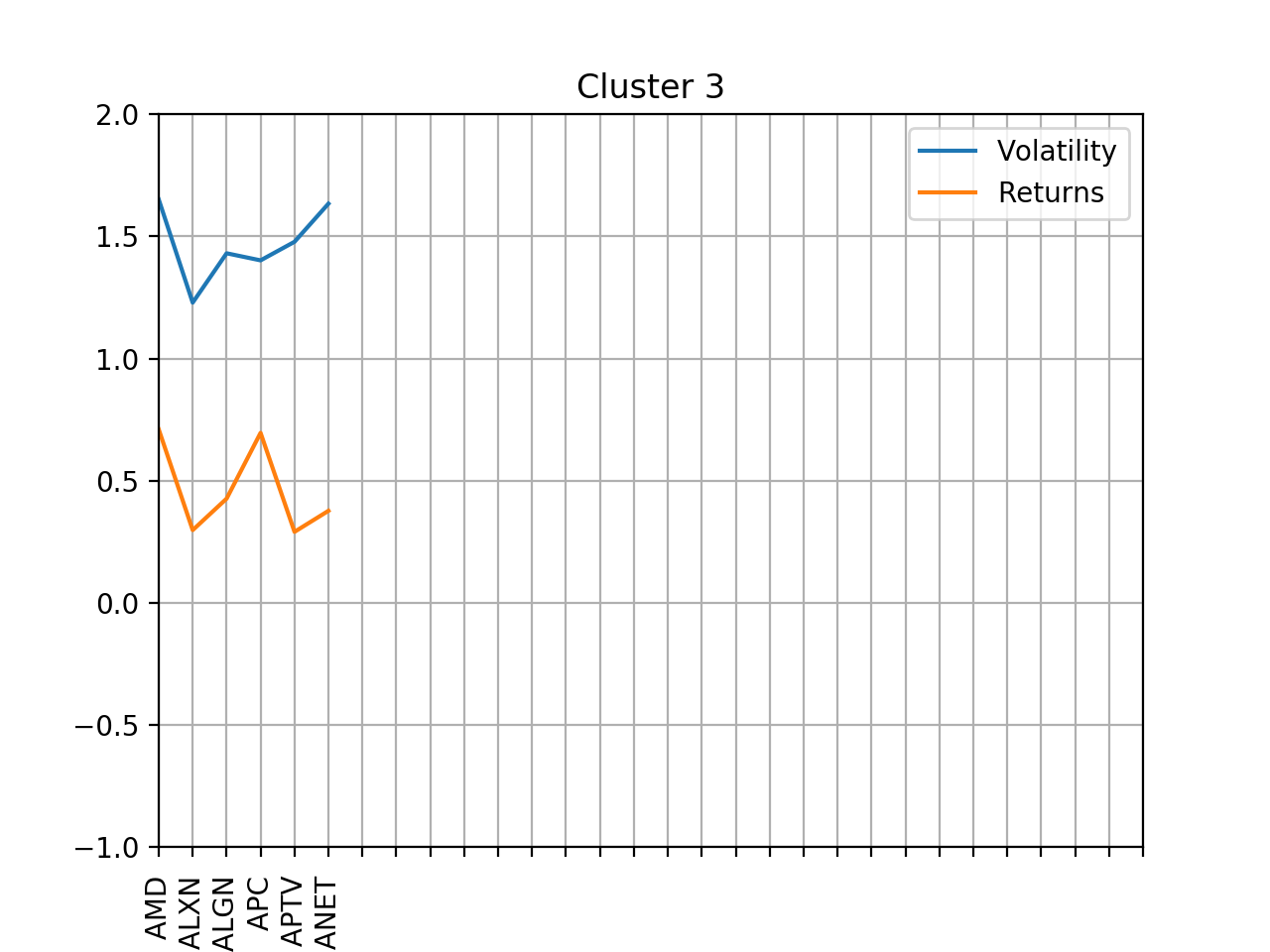}
        \caption{KMeans clustering: The range of volatility and returns for Cluster "3"}
        \label{fig:kmeans-3-VR}
    \end{minipage}
\end{figure*}

Figures \ref{fig:kmeans-0-VR} - \ref{fig:kmeans-3-VR}
illustrate the range of volatility and return values of stock indices that are clustered together. The stock indices are clustered with respect to two descriptive variables $<volatility, return>$. 
As a result, the trends of these two variables is similar to the other stock indices clustered in the same group. The figures  visualize the range of volatility and returns computed for each member of clusters produces by KMeans clustering for the period of January 1, 2019 to April 15, 2019.  

\subsection{Encoder-Decoder: Parameters Trained} 
Table \ref{tab:EncoderDecoder} lists the number of layers, the output shape (i.e., the number of nodes or neurons) of each layer along with the number of parameters estimated at each layer.

\begin{table}[t]
\caption{The input and output shapes along with the number of parameters trained.}
\label{tab:EncoderDecoder}
\begin{center}
%\begin{tabular}{p{7.5cm}}
\begin{tabular}{|c|c|c|r|}
\hline
& {\bf Layer (Type)} & {\bf Output Shape} & {\bf Parameter\#} \\
\hline
1& Input 1 (Input Layer) & (None, 2) & 0 \\
2& dense\_1 (Dense) & (None, 100) & 300 \\
3& dense\_2 (Dense) & (None, 50) & 5,050 \\
4& dense\_3 (Dense) & (None, 20) & 1,020 \\
5& dense\_4 (Dense) & (None, 4) & 84 \\
6& dense\_5 (Dense) & (None, 20) & 100 \\
7& dense\_6 (Dense) & (None, 50) & 1,050 \\
8& dense\_7 (Dense) & (None, 100) & 5,100 \\
9& dense\_8 (Dense) & (None, 1) & 101 \\ 
\hline
\multicolumn{3}{|c|}{\bf Total Trainable Parameters} & 12,805 \\
\hline
\end{tabular}
\end{center}
\end{table}

As highlighted earlier, the authoencoder takes as input a feature vector of size 2 (i.e., its shape), and then propagate the input to the internal layers devised for the encoder and decoder parts. at the level $f$ dense\_4 the shape is in the form of 4, the number of desired clusters. the exact number of layers and nodes are built in a reverse order and eventually an output shape with one column is produced. The total number of trained parameters is $12,805$, which implies creating a fully connected network. 

\begin{figure}[h]%[!t]
  \includegraphics[width=\linewidth]{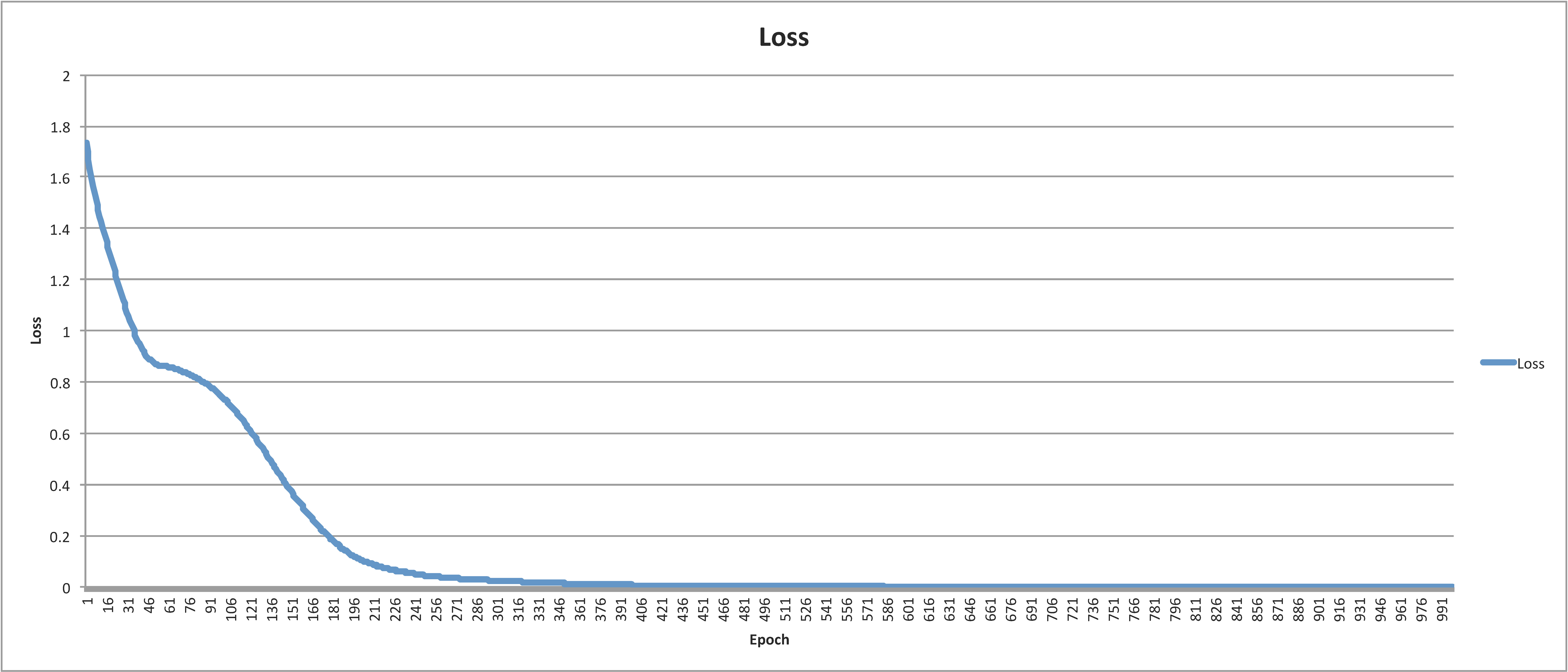}
  \caption{Loss vs. epochs.}
  \label{fig:epochs}
  %\vspace{-0.15in}
\end{figure}

We trained the model with different number of repetition (i.e., epoch) in order to understand the performance of estimating the parameter values in details. We repeated the training model for 1000 epochs. Figure \ref{fig:epochs} illustrates the relationship between number of epochs and the error (i.e., loss). As the figure indicates, the loss value is approximately zero when the number of epochs is greater than 316.

We kept the number of epochs as 1000, even though 316 epochs were sufficient. Once the network is trained using the training dataset, it is given the test dataset to predict the labels. The prediction is in the form of a numeral value that needs to be rounded. The computed numerical output of the program and the rounded with absolute $d$ values along with the actual label for the test data set are reported in Table \ref{tab:prediction}.

\begin{table*}[t]
\caption{Numerical prediction of time series' cluster labels.}
\label{tab:prediction}
\begin{center}
%\begin{tabular}{p{7.5cm}}
\begin{tabular}{|c|c|r|r|r|c|c|c|}
\hline
& \multicolumn{1}{|c|}{\bf Index} & \multicolumn{1}{|c|}{\bf Volatility} & \multicolumn{1}{|c|}{\bf Returns} & \multicolumn{1}{|c|}{\bf Exact Output}  & \multicolumn{1}{|c|}{\bf Absolute Rounded} & \multicolumn{1}{|c|}{\bf KMeans} & \multicolumn{1}{|c|}{\bf Missed}  \\
& & & & \multicolumn{1}{|c|}{\bf Label Prediction} &  \multicolumn{1}{|c|}{\bf Prediction} & \multicolumn{1}{|c|}{\bf Cluster } & \multicolumn{1}{|c|}{\bf Labeled} \\
& & & & \multicolumn{1}{|c|}{\bf ($r$)} &  \multicolumn{1}{|c|}{\bf (Predicted Label)} & \multicolumn{1}{|c|}{\bf Label} & \multicolumn{1}{|c|}{\bf } \\
\hline
1 & ADS & 0.317 & 0.612 & 7.2130698e-01 & 1. & 1 & \\
2 & MMM & 0.201 & 0.512 & 9.9868220e-01 & 1. & 1 & \\
3 & AAPL & 0.308 & 0.894 & -1.8404983e-04 & 0. & 0 & \\
4 & ACN & 0.179 & 0.909 & -1.4865603e-03 & 0. & 0 & \\
5 & ANET & 0.377 & 1.634 & 3.0150795e+00 & 3. & 3 & \\
{\bf \color{red} 6} & {\bf \color{red}  ALXN} & {\bf \color{red}  0.298} & {\bf \color{red}  1.229} & {\bf \color{red}  2.4604988e+00} & {\bf \color{red}  2.} & {\bf \color{red}  3} & {\bf \color{red}  X}\\
7 & AMG & 0.296 & 0.537 & 9.9893832e-01 & 1. & 1 & \\
8 & AIG & 0.276 & 0.617 & 8.0927080e-01 & 1. & 1 & \\
9 & AON & 0.254 & 0.766 & 4.7755931e-03 & 0. & 0 & \\
10 & A & 0.200 & 0.781 & 2.7296934e-03 & 0. & 0 & \\
11 & AEP & 0.125 & 0.553 & 9.9907714e-01 & 1. & 1 & \\
12 & AES & 0.172 & 0.909 & -1.5225317e-03 & 0. & 0 & \\
13 & BK & 0.183 & 0.417 & 9.9732614e-01 & 1. & 1 & \\
14 & ATVI & 0.460 & 0.154 & 1.9892873e+00 & 2. & 2 & \\
15 & AVB & 0.102 & 0.708 & 7.7624805e-02 & 0. & 0 & \\
16 & AAL & 0.379 & 0.317 & 1.0931786e+00 & 1. & 1 & \\
17 & T & 0.182 & 0.443 & 9.9788338e-01 & 1. & 1 & \\
18 & AWK & 0.123 & 0.600 & 9.9946052e-01 & 1. & 1 & \\
19 & ATO & 0.140 & 0.457 & 9.9807245e-01 & 1. & 1 & \\
{\bf \color{red}  20} & {\bf \color{red}  APA} & {\bf \color{red}  0.336} & {\bf \color{red}  1.157} & {\bf \color{red}  2.0670881e+00} & {\bf \color{red}  2.} & {\bf \color{red}  0} & {\bf \color{red}  X} \\
21 & ALB & 0.333 & 0.317 & 9.9551427e-01 & 1. & 1 & \\
22 & AMT & 0.123 & 0.866 & -1.1737701e-03 & 0. & 0 & \\
23 & ADI & 0.285 & 1.085 & 1.2981926e-01 & 0. & 0 & \\
{\bf \color{red}  24} & {\bf \color{red}  AMZN} & {\bf \color{red}  0.284} & {\bf \color{red}  0.689} & {\bf \color{red}  3.1280313e-03} & {\bf \color{red}  0.} & {\bf \color{red}  1} & {\bf \color{red}  X}\\
\hline
\end{tabular}
\end{center}
\end{table*}

The total number of data set was 70 (i.e., the time series data 70 stock indices were captured), of which 46 time series data were considered for training the network, and the remaining data set (i.e., 24) was used for testing the model. As Table \ref{tab:prediction} lists, the network was able to predict the cluster label of 21 out of 24 test data correctly achieving an accuracy of $87.5$ in prediction. The miss-classified stock indices are ALXN, APA, and AMZN which are colored in red in the figures. 

\begin{figure*}[t]
    \centering
    \begin{minipage}{.45\textwidth}
        \centering
        \includegraphics[width=0.9\linewidth]{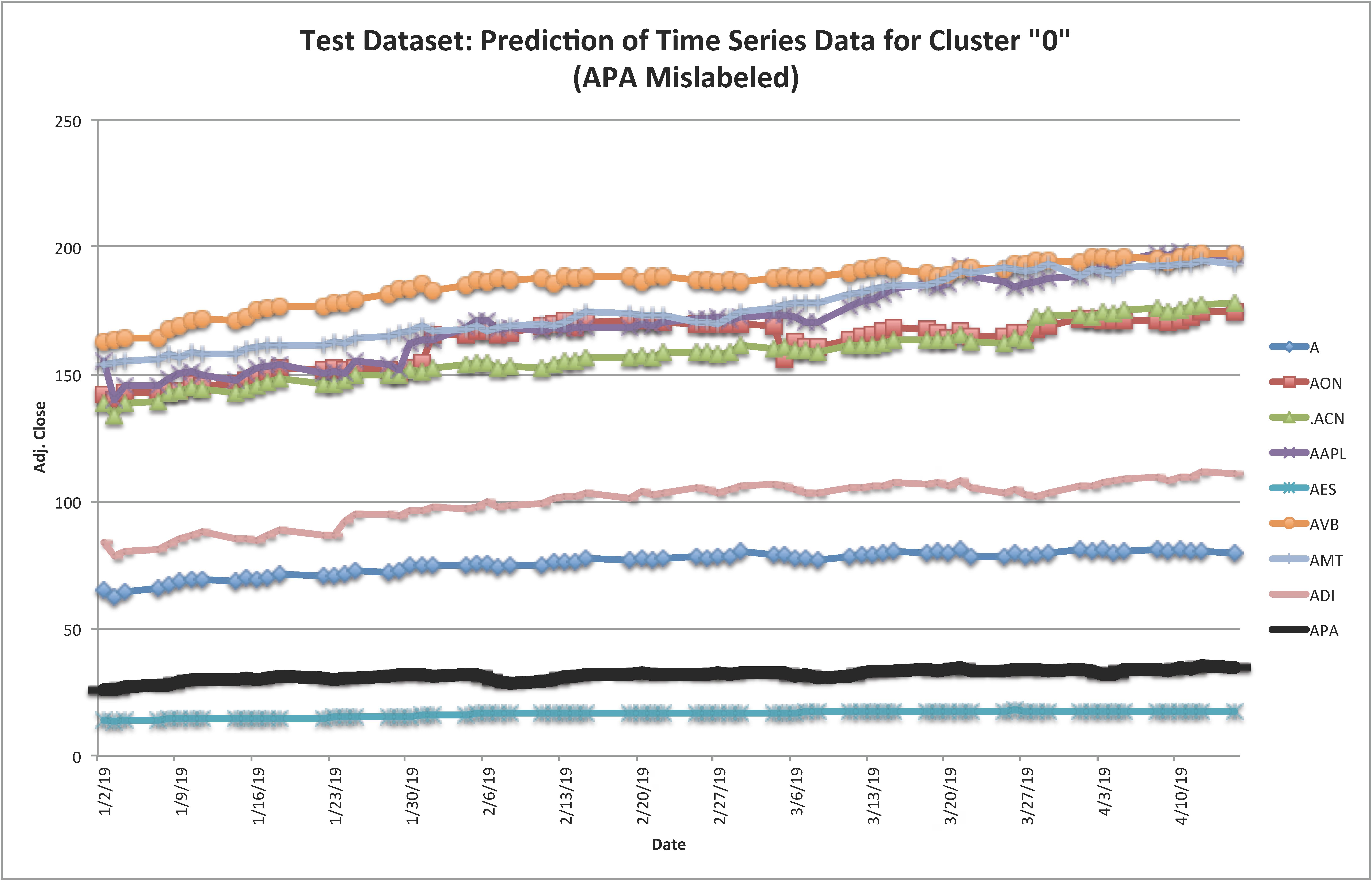}
        \caption{Prediction of time series data for Cluster "0"}
        \label{fig:prediction-0}
    \end{minipage}%
    \begin{minipage}{0.45\textwidth}
        \centering
        \includegraphics[width=0.9\linewidth]{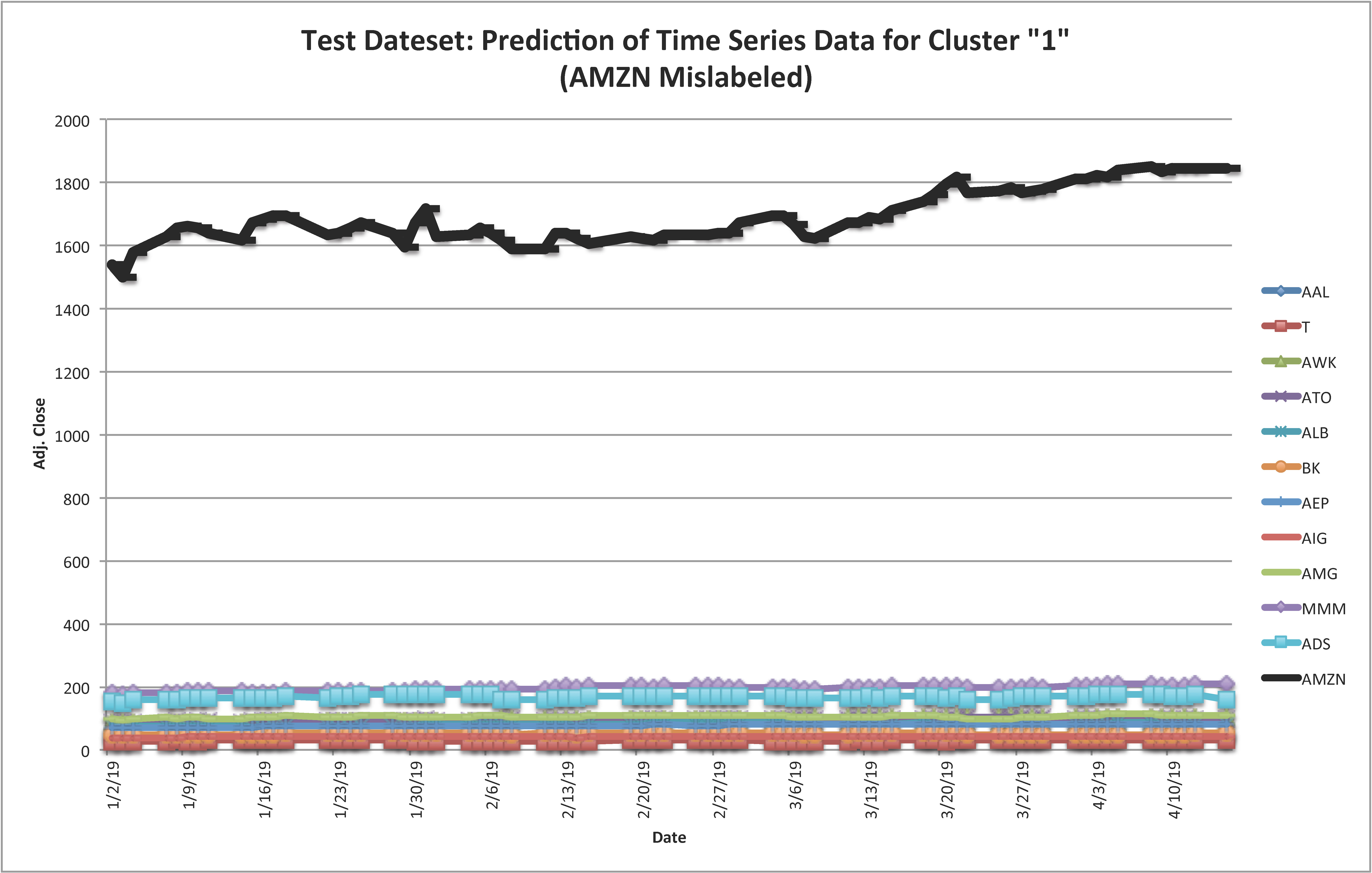}
        \caption{Prediction of time series data for Cluster "1"}
        \label{fig:prediction-1}
    \end{minipage}
        \begin{minipage}{0.45\textwidth}
        \centering
        \includegraphics[width=0.9\linewidth]{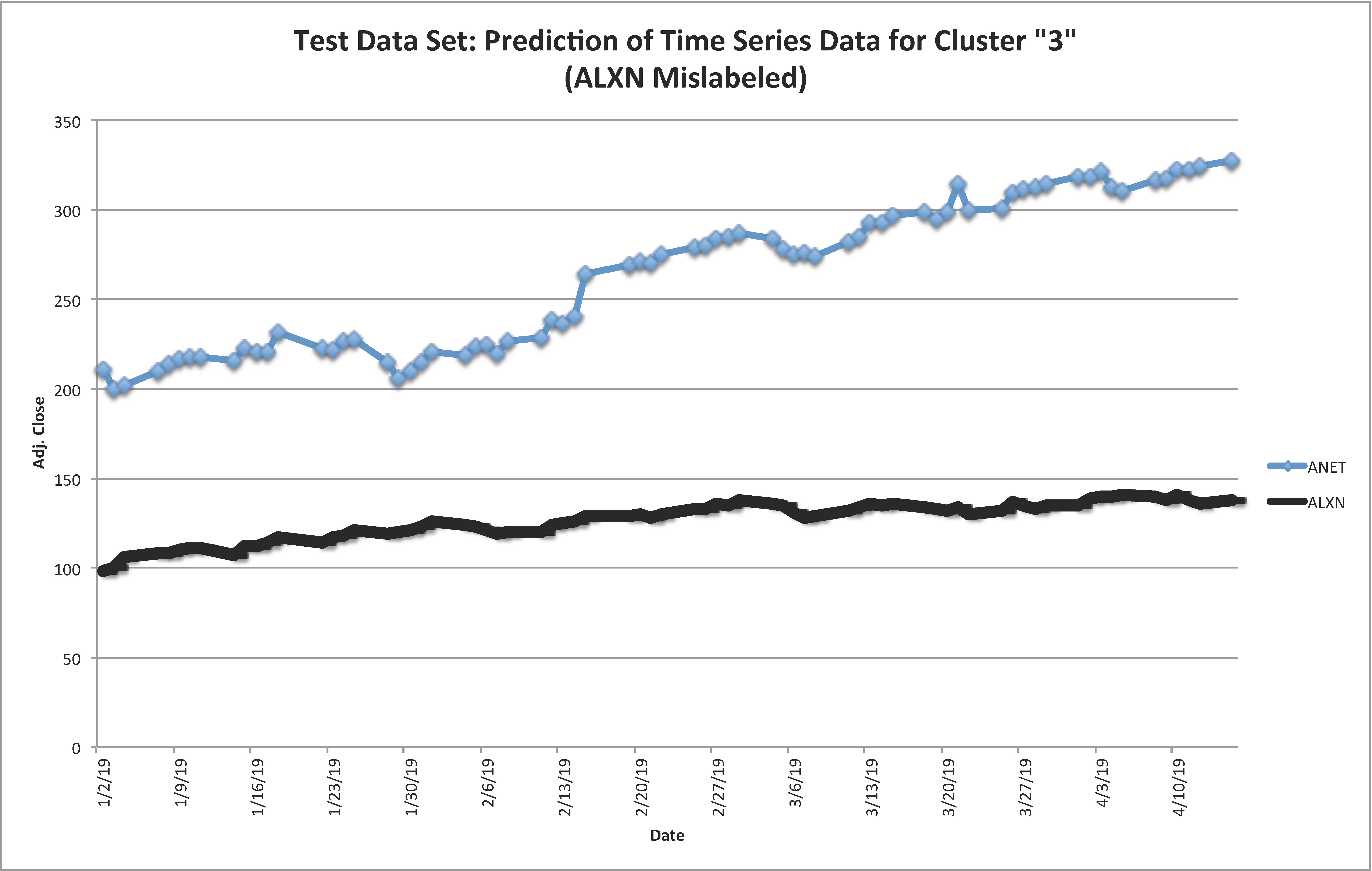}
        \caption{Prediction of time series data for Cluster "3"}
        \label{fig:prediction-2}
    \end{minipage}
        \begin{minipage}{0.45\textwidth}
        \centering
        \includegraphics[width=0.9\linewidth]{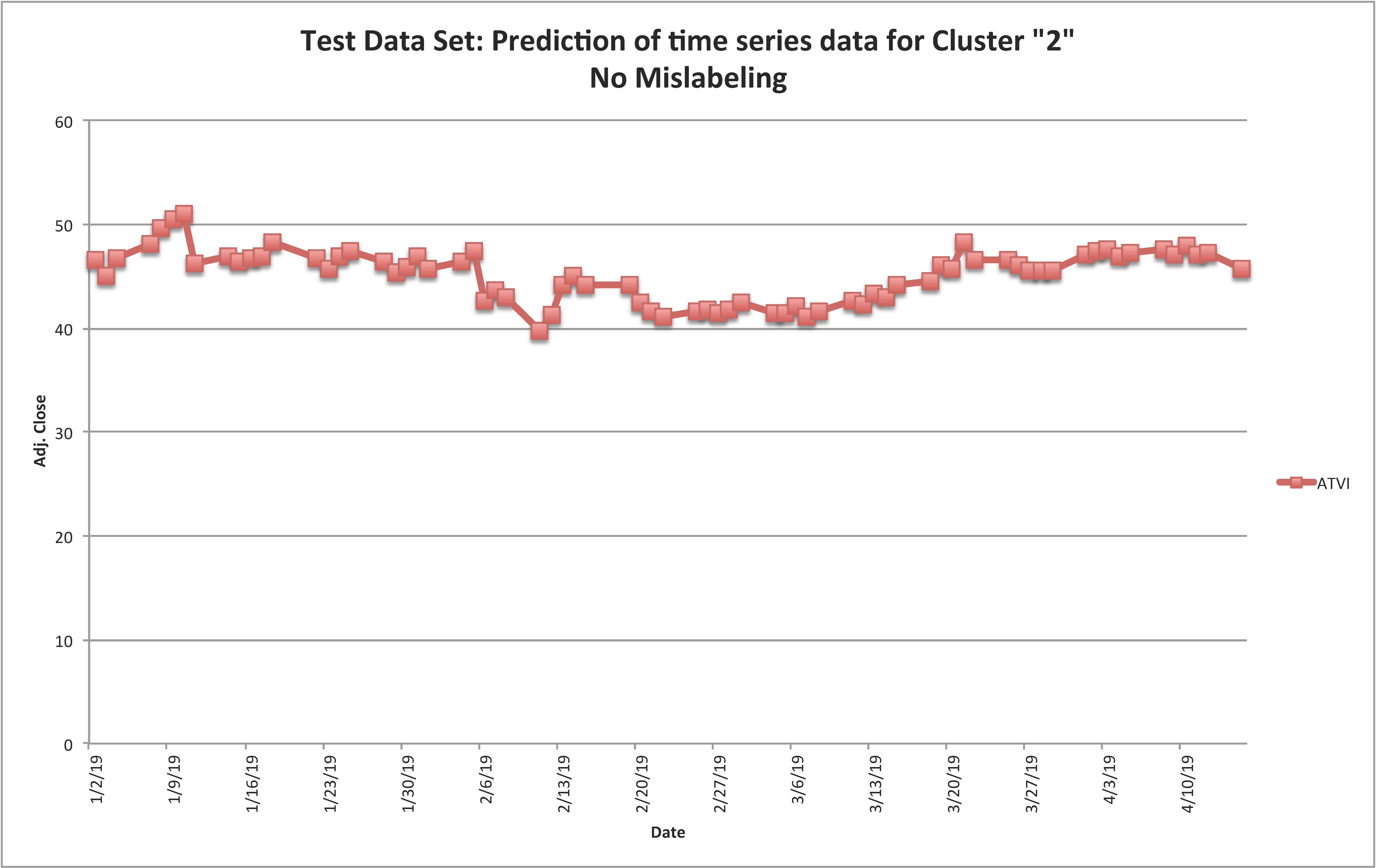}
        \caption{Prediction of time series data for Cluster "2"}
        \label{fig:prediction-3}
    \end{minipage}
\end{figure*}

To help realize the results of the autoencoder-based deep learning time series classification model, we visualize the results. Figures \ref{fig:prediction-0} - \ref{fig:prediction-3} show the results of the prediction of cluster's label for the test set in which a time series with black color (i.e., except Cluster ``2'') show the miss-classifications performed by the prediction. The prediction results in three instances of mislabeling colored in black in the figures. 

\begin{figure*}[t]
    \centering
    \begin{minipage}{.45\textwidth}
        \centering
        \includegraphics[width=0.9\linewidth]{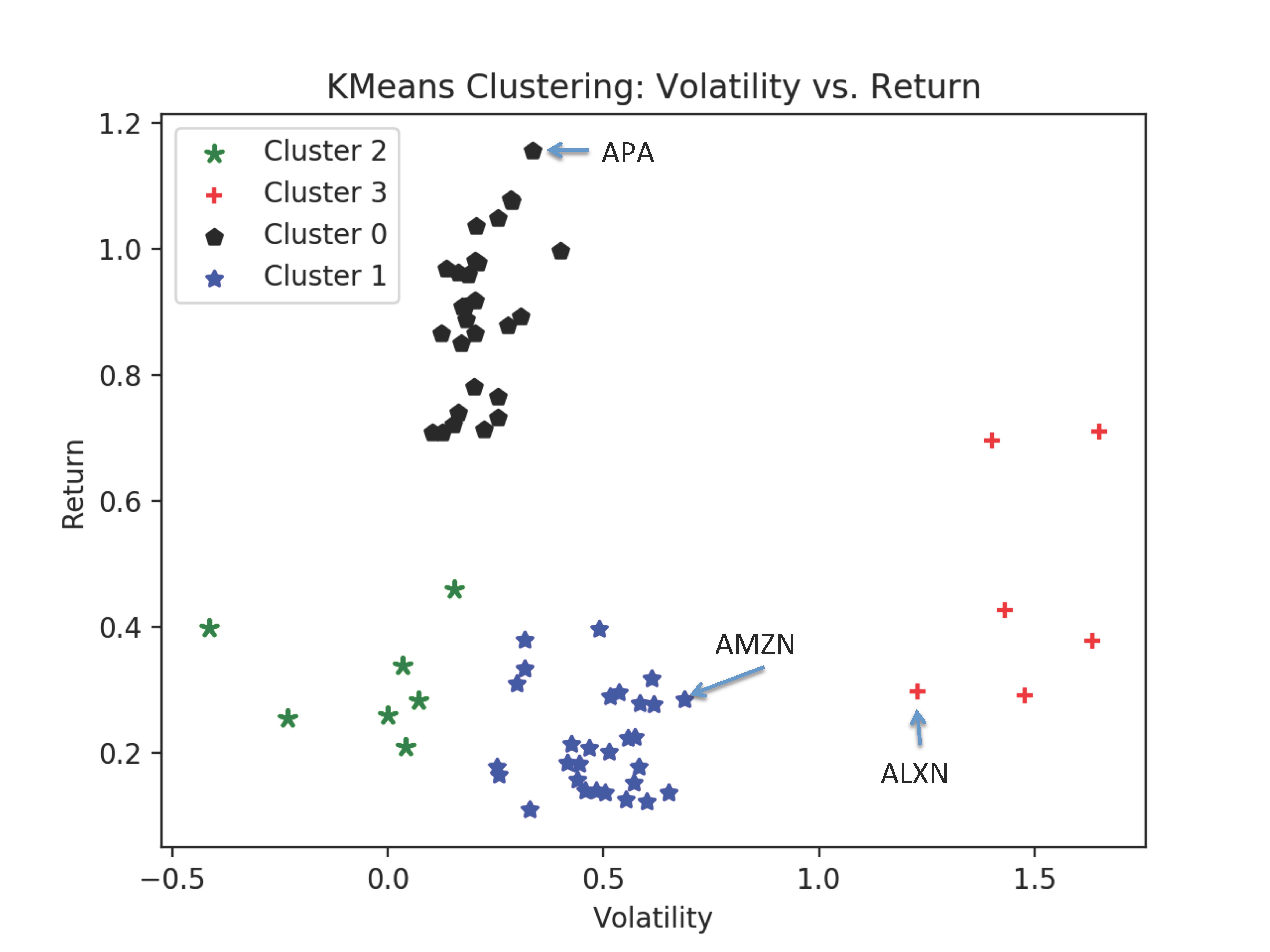}
        \caption{KMeans Clustering}
        \label{fig:KMeansClustering}
    \end{minipage}%
    \begin{minipage}{0.45\textwidth}
        \centering
        \includegraphics[width=0.9\linewidth]{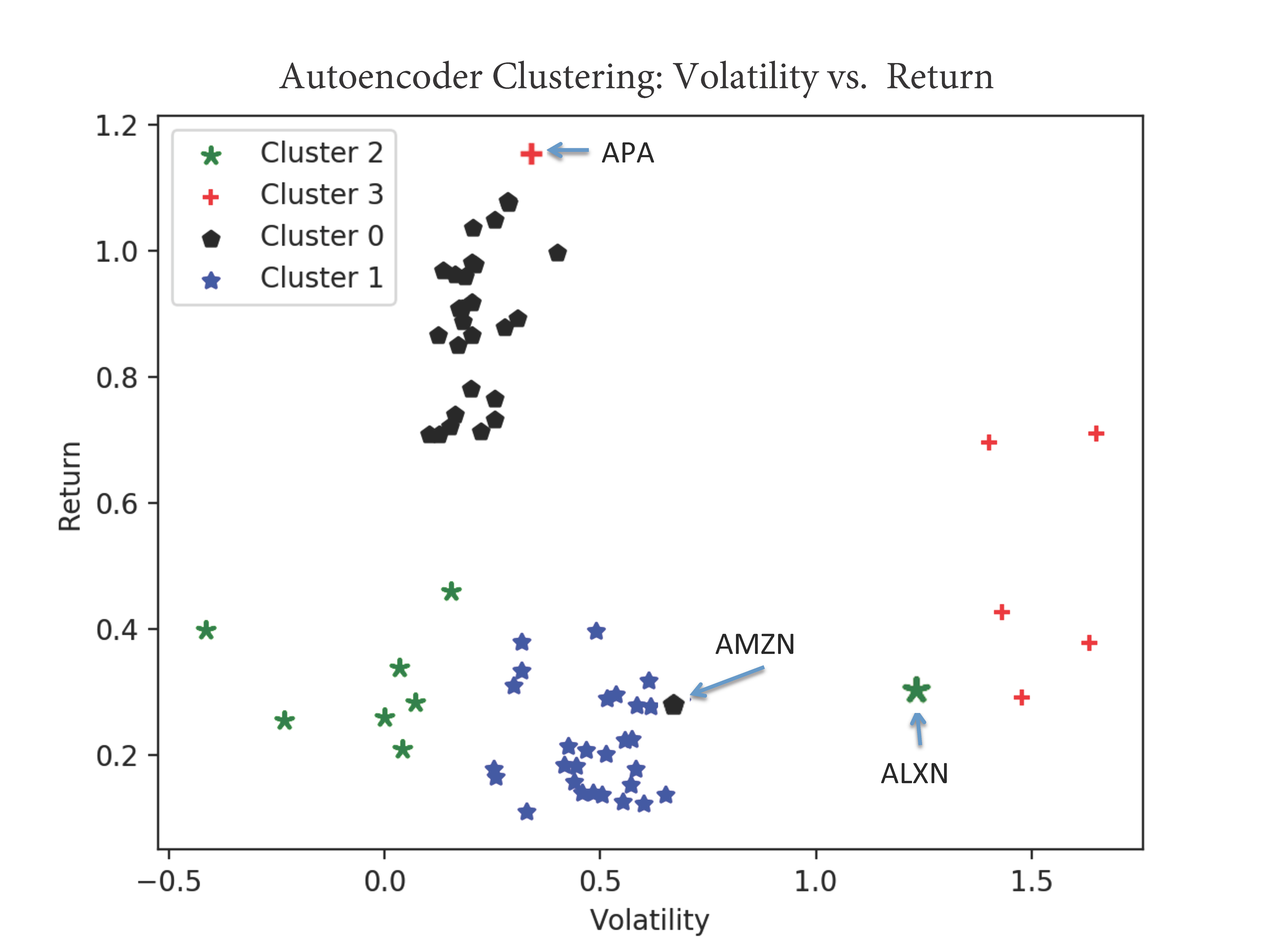}
        \caption{Autoencoder Clustering}
        \label{fig:DLClustering}
    \end{minipage}
\end{figure*}

\subsection{KMenas vs. Deep Learning-based Clustering}

The  investigation  of why ALXN, APA, and AMLN are miss-classified reveals interesting findings. The clustering performed by these two are illustrted in Figures \ref{fig:KMeansClustering} and \ref{fig:DLClustering}. By refering to Table \ref{tab:KMeans}, in which the results of KMeans clustering are reported, we observe that:
\begin{itemize}
\renewcommand\labelitemi{--}
    \item the feature vector of AMZN is $<0.284, 0.689>$. A comparison of the feature vector for AMZN with those clustered together in Cluster "1" by conventional KMeans show that the return value for AMZN is on the upper bound of the return values clustered in Cluster "1" (i.e., it is the max value for the return).  
    \item A similar finding is observable for APA which is clustered by conventional KMeans in Cluster "0" with $0.336, 1.157$ (Table \ref{tab:KMeans}). Similarly, both volatility and return values calculated for APA are on the upper bounds of the volatility and return values calculated for stock indices cluster together in Cluster "0".  
    \item similarly, the feature vector calculated for ALXN is $0.298, 1.229$ which both are on the lower bounds of the feature vectors for volatility and returns clustered together in Cluster "2". 
\end{itemize}

The findings indicate that, these three stock indices (i.e., ALXN, APA, and AMZN) are on the border line of clusters (See Figures \ref{fig:KMeansClustering} and \ref{fig:DLClustering}). Even though the figures may imply that the clustering produced by KMeans has been performed reasonably well, it may also indicate that clustering performed by the autoencoder might have taken into account some other hidden factors. Hence, since deep learning-based approach is discovering and taking into account more hidden features among these two values (i.e., volatility and return), the clustering performed by the autoencoder is actually providing more insights about these stock indices and their relationships. More precisely, it might indicate that there might be some other hidden features discovered by the autoencoder that are missed and not formulated by conventional KMeans clustering algorithms.

\section{Conclusions and Future Work}
\label{sec:conclusions}
Time series are complex data with characterized by several features. These features are often utilized by time series data analyses in order to understand the behavior and nature of the underlying application domains and further for prediction of the trend of data. Existing statistical techniques such as ARIMA (autoregressive integrated moving average) and regressions are capable of linearly predict the trends of the data. These conventional statistical techniques utilize known features such as seasonal effects, cycles, and trend in order to build prediction models. Although these features and techniques have acceptable accuracy, their performance might be deteriorated mainly due to the existence of some other and hidden features which are not part of the prediction or classification models. 

Deep learning-based techniques are emerging approaches to data sciences and analyses. These techniques are capable of detecting hidden features of data and thus take into account these features when building prediction models. In particular, these techniques are expected to outperform traditional statistical data analysis in the context of time series, due to the additional complexity added by the time factor to the data. 

This paper introduces a deep learning-based approach to model the time series clustering problem, in which the given time series data are clustered into groups with respect to some features. The time series clustering solution presented in this paper first generates labels for time series data using KMeans clustering and thus enabling supervised learning. Once the cluster labels are generated, they are given to an encode-decoder-based deep learning neural network in order to build a clustering and prediction model. The most important advantage of building such as neural network is that it models hidden features and takes into account such features into the prediction. The case study conducted in the context of the financial time series data shows the accuracy of \%87.5 in clustering such data. More importantly, we observed that the deep learning-based model outperforms the conventional KMeans clustering. 

The application of deep learning approaches to time series analysis and in particular financial time series data is in its early stages. Several other classical problems in time series analysis can be formulated using deep learning techniques such as shock and anomaly detection, seasonal effects as well as clustering and prediction at different levels of data abstractions. Neural network-based techniques such as Long Short Term Memory (LSTM) \cite{ICMLA, COMPSACNeda, BigData2019BiLSTM, siami2019comparative, Abrietal2019}, Generative Adversarial Networks (GANS), and many others need to be further explored for formulating classical problems in time series data analysis.

% or


\begin{comment}
\appendices
\section{XXX}
Appendix one text goes here.

\section{}
Appendix two text goes here.

\section*{Acknowledgment}

The authors would like to thank...
\end{comment}

\section*{Conflict of Interests Statement}
The authors do not have any conflict of interests to disclose. 

\section*{Funding Statement}
This work is partially funded by the support from National Science Foundation under the grant numbers 1564293, 1723765, and 1821560.

\ifCLASSOPTIONcaptionsoff
  \newpage
\fi



%\bibliographystyle{IEEEtran}
%\bibliographystyle{IEEEtran}
\bibliographystyle{ieeetr}
\bibliography{sample-base}{}

\begin{thebibliography}{10}

\bibitem{chen2007spade}
Y.~Chen, M.~A. Nascimento, B.~C. Ooi, and A.~K. Tung, ``Spade: On shape-based
  pattern detection in streaming time series,'' in {\em 2007 IEEE 23rd
  International Conference on Data Engineering}, pp.~786--795, IEEE, 2007.

\bibitem{ding2008querying}
H.~Ding, G.~Trajcevski, P.~Scheuermann, X.~Wang, and E.~Keogh, ``Querying and
  mining of time series data: experimental comparison of representations and
  distance measures,'' {\em Proceedings of the VLDB Endowment}, vol.~1, no.~2,
  pp.~1542--1552, 2008.

\bibitem{stefan2013move}
A.~Stefan, V.~Athitsos, and G.~Das, ``The move-split-merge metric for time
  series,'' {\em IEEE transactions on Knowledge and Data Engineering}, vol.~25,
  no.~6, pp.~1425--1438, 2013.

\bibitem{wang2013experimental}
X.~Wang, A.~Mueen, H.~Ding, G.~Trajcevski, P.~Scheuermann, and E.~Keogh,
  ``Experimental comparison of representation methods and distance measures for
  time series data,'' {\em Data Mining and Knowledge Discovery}, vol.~26,
  no.~2, pp.~275--309, 2013.

\bibitem{aghabozorgi2015time}
S.~Aghabozorgi, A.~S. Shirkhorshidi, and T.~Y. Wah, ``Time-series clustering--a
  decade review,'' {\em Information Systems}, vol.~53, pp.~16--38, 2015.

\bibitem{keogh2005clustering}
E.~Keogh and J.~Lin, ``Clustering of time-series subsequences is meaningless:
  implications for previous and future research,'' {\em Knowledge and
  information systems}, vol.~8, no.~2, pp.~154--177, 2005.

\bibitem{liao2005clustering}
T.~W. Liao, ``Clustering of time series data—a survey,'' {\em Pattern
  recognition}, vol.~38, no.~11, pp.~1857--1874, 2005.

\bibitem{tavakoli2019client}
N.~Tavakoli, D.~Dai, and Y.~Chen, ``Client-side straggler-aware i/o scheduler
  for object-based parallel file systems,'' {\em Parallel Computing}, vol.~82,
  pp.~3--18, 2019.

\bibitem{tavakoli2016log}
N.~Tavakoli, D.~Dai, and Y.~Chen, ``Log-assisted straggler-aware i/o scheduler
  for high-end computing,'' in {\em 2016 45th International Conference on
  Parallel Processing Workshops (ICPPW)}, pp.~181--189, IEEE, 2016.

\bibitem{vlachos2004indexing}
M.~Vlachos and D.~Gunopulos, ``Indexing time series under condition of noise.
  data mining in time series database: Series in machine perception and
  artificial intelligence,'' 2004.

\bibitem{ratanamahatana2005novel}
C.~Ratanamahatana, E.~Keogh, A.~J. Bagnall, and S.~Lonardi, ``A novel bit level
  time series representation with implication of similarity search and
  clustering,'' in {\em Pacific-Asia Conference on Knowledge Discovery and Data
  Mining}, pp.~771--777, Springer, 2005.

\bibitem{keogh2003need}
E.~Keogh and S.~Kasetty, ``On the need for time series data mining benchmarks:
  a survey and empirical demonstration,'' {\em Data Mining and knowledge
  discovery}, vol.~7, no.~4, pp.~349--371, 2003.

\bibitem{lin2007experiencing}
J.~Lin, E.~Keogh, L.~Wei, and S.~Lonardi, ``Experiencing sax: a novel symbolic
  representation of time series,'' {\em Data Mining and knowledge discovery},
  vol.~15, no.~2, pp.~107--144, 2007.

\bibitem{popivanov2002similarity}
I.~Popivanov and R.~J. Miller, ``Similarity search over time-series data using
  wavelets,'' in {\em Proceedings 18th international conference on data
  engineering}, pp.~212--221, IEEE, 2002.

\bibitem{lin2003symbolic}
J.~Lin, E.~Keogh, S.~Lonardi, and B.~Chiu, ``A symbolic representation of time
  series, with implications for streaming algorithms,'' in {\em Proceedings of
  the 8th ACM SIGMOD workshop on Research issues in data mining and knowledge
  discovery}, pp.~2--11, ACM, 2003.

\bibitem{bagnall2006bit}
A.~Bagnall, E.~Keogh, S.~Lonardi, G.~Janacek, {\em et~al.}, ``A bit level
  representation for time series data mining with shape based similarity,''
  {\em Data Mining and Knowledge Discovery}, vol.~13, no.~1, pp.~11--40, 2006.

\bibitem{shieh2008sax}
J.~Shieh and E.~Keogh, ``i sax: indexing and mining terabyte sized time
  series,'' in {\em Proceedings of the 14th ACM SIGKDD international conference
  on Knowledge discovery and data mining}, pp.~623--631, ACM, 2008.

\bibitem{minnen2007discovering}
D.~Minnen, C.~L. Isbell, I.~Essa, and T.~Starner, ``Discovering multivariate
  motifs using subsequence density estimation and greedy mixture learning,'' in
  {\em Proceedings of the National Conference on Artificial Intelligence},
  vol.~22, p.~615, Menlo Park, CA; Cambridge, MA; London; AAAI Press; MIT
  Press; 1999, 2007.

\bibitem{kumar2005time}
N.~Kumar, V.~N. Lolla, E.~Keogh, S.~Lonardi, C.~A. Ratanamahatana, and L.~Wei,
  ``Time-series bitmaps: a practical visualization tool for working with large
  time series databases,'' in {\em Proceedings of the 2005 SIAM international
  conference on data mining}, pp.~531--535, SIAM, 2005.

\bibitem{kalpakis2001distance}
K.~Kalpakis, D.~Gada, and V.~Puttagunta, ``Distance measures for effective
  clustering of arima time-series,'' in {\em Proceedings 2001 IEEE
  international conference on data mining}, pp.~273--280, IEEE, 2001.

\bibitem{bagnall2005clustering}
A.~Bagnall and G.~Janacek, ``Clustering time series with clipped data,'' {\em
  Machine Learning}, vol.~58, no.~2-3, pp.~151--178, 2005.

\bibitem{chu2002iterative}
S.~Chu, E.~Keogh, D.~Hart, and M.~Pazzani, ``Iterative deepening dynamic time
  warping for time series,'' in {\em Proceedings of the 2002 SIAM International
  Conference on Data Mining}, pp.~195--212, SIAM, 2002.

\bibitem{wang2006characteristic}
X.~Wang, K.~Smith, and R.~Hyndman, ``Characteristic-based clustering for time
  series data,'' {\em Data mining and knowledge Discovery}, vol.~13, no.~3,
  pp.~335--364, 2006.

\bibitem{kaufman2009finding}
L.~Kaufman and P.~J. Rousseeuw, {\em Finding groups in data: an introduction to
  cluster analysis}, vol.~344.
\newblock John Wiley \& Sons, 2009.

\bibitem{keogh1998enhanced}
E.~J. Keogh and M.~J. Pazzani, ``An enhanced representation of time series
  which allows fast and accurate classification, clustering and relevance
  feedback.,'' in {\em Kdd}, vol.~98, pp.~239--243, 1998.

\bibitem{gupta1996nonlinear}
L.~Gupta, D.~L. Molfese, R.~Tammana, and P.~G. Simos, ``Nonlinear alignment and
  averaging for estimating the evoked potential,'' {\em IEEE Transactions on
  Biomedical Engineering}, vol.~43, no.~4, pp.~348--356, 1996.

\bibitem{hautamaki2008time}
V.~Hautamaki, P.~Nykanen, and P.~Franti, ``Time-series clustering by
  approximate prototypes,'' in {\em 2008 19th International Conference on
  Pattern Recognition}, pp.~1--4, IEEE, 2008.

\bibitem{macqueen1967some}
J.~MacQueen {\em et~al.}, ``Some methods for classification and analysis of
  multivariate observations,'' in {\em Proceedings of the fifth Berkeley
  symposium on mathematical statistics and probability}, vol.~1, pp.~281--297,
  Oakland, CA, USA, 1967.

\bibitem{ester1996density}
M.~Ester, H.-P. Kriegel, J.~Sander, X.~Xu, {\em et~al.}, ``A density-based
  algorithm for discovering clusters in large spatial databases with noise.,''
  in {\em Kdd}, vol.~96, pp.~226--231, 1996.

\bibitem{wang1997sting}
W.~Wang, J.~Yang, R.~Muntz, {\em et~al.}, ``Sting: A statistical information
  grid approach to spatial data mining,'' in {\em VLDB}, vol.~97, pp.~186--195,
  1997.

\bibitem{sheikholeslami1998wavecluster}
G.~Sheikholeslami, S.~Chatterjee, and A.~Zhang, ``Wavecluster: A
  multi-resolution clustering approach for very large spatial databases,'' in
  {\em VLDB}, vol.~98, pp.~428--439, 1998.

\bibitem{fu2001pattern}
T.-c. Fu, F.-l. Chung, V.~Ng, and R.~Luk, ``Pattern discovery from stock time
  series using self-organizing maps,'' in {\em Workshop Notes of KDD2001
  Workshop on Temporal Data Mining}, pp.~26--29, 2001.

\bibitem{aghabozorgi2014hybrid}
S.~Aghabozorgi, T.~Ying~Wah, T.~Herawan, H.~A. Jalab, M.~A. Shaygan, and
  A.~Jalali, ``A hybrid algorithm for clustering of time series data based on
  affinity search technique,'' {\em The Scientific World Journal}, vol.~2014,
  2014.

\bibitem{lai2010novel}
C.-P. Lai, P.-C. Chung, and V.~S. Tseng, ``A novel two-level clustering method
  for time series data analysis,'' {\em Expert Systems with Applications},
  vol.~37, no.~9, pp.~6319--6326, 2010.

\bibitem{Sewell}
M.~Sewell, ``Characterization of financial time series,'' 2011.

\bibitem{Mamtha}
D.~Mamtha and K.~S. Srinivasan, ``Stock market volatility: Conceptual
  perspective through literature survey,'' {\em Mediterranean Journal of Social
  Sciences}, vol.~7, no.~1, pp.~208 -- 212, 2016.

\bibitem{Bakaert}
G.~Bakaert and G.~Wu, ``Asymmetric volatility and risk in equity markets,''
  {\em Review of financial Studies}, vol.~13, no.~1, pp.~1 -- 42, 2000.

\bibitem{Whitelaw}
R.~Whitelaw, ``Stock market risk and return: An empirical equilibrium
  approach,'' {\em Review of financial Studies}, vol.~13, no.~3, pp.~521 --
  547, 2000.

\bibitem{Shawky}
H.~A. Shawky and A.~Marathe, ``Expected stock returns and volatility in a two
  regime market,'' {\em The Journal of Economics and Business}, vol.~47, no.~5,
  pp.~409 -- 422, 1995.

\bibitem{Chung}
K.~Chung and C.~Chuwongananat, ``Market volatility and stock returns: The role
  of liquidity providers,'' {\em Journal of Financial Markets}, vol.~37, pp.~17
  -- 34, 2018.

\bibitem{Sen}
Sen and Bandhopadhyay, ``On the return and volatility spillover between us and
  indian stock market,'' {\em International Journal of Financial Management},
  vol.~1, no.~3, 2012.

\bibitem{Hubens}
N.~Hubens, ``Deep inside: Autoencoders,'' 2019.

\bibitem{ICMLA}
S.~Siami{-}Namini, N.~Tavakoli, and A.~S. Namin, ``A comparison of {ARIMA} and
  {LSTM} in forecasting time series,'' in {\em 17th {IEEE} International
  Conference on Machine Learning and Applications, {ICMLA} 2018, Orlando, FL,
  USA, December 17-20, 2018}, pp.~1394--1401, 2018.

\bibitem{COMPSACNeda}
N.~Tavakoli, ``Modeling genome data using bidirectional {LSTM},'' in {\em The
  1st IEEE International Workshop on Deep Analysis of Data-Driven Applications
  (DADA) in conjunction with COMPSAC}, 2019.

\bibitem{BigData2019BiLSTM}
S.~Siami-Namini, N.~Tavakoli, and A.~S. Namin, ``The performance of {LSTM} and
  {BiLSTM} in forecasting time series,'' in {\em IEEE Big Data, Los Angeles,
  California, USA}, 2019.

\bibitem{siami2019comparative}
S.~Siami-Namini, N.~Tavakoli, and A.~S. Namin, ``A comparative analysis of
  forecasting financial time series using arima, lstm, and bilstm,'' {\em arXiv
  preprint arXiv:1911.09512}, 2019.

\bibitem{Abrietal2019}
F.~Abri, S.~Siami-Namini, M.~A. Khanghah, F.~M. Soltani, and A.~S. Namin, ``Can
  machine/deep learning classifiers detect zero-day malware with high
  accuracy?,'' in {\em IEEE Big Data, Los Angeles, California, USA}, 2019.

\end{thebibliography}

\end{document}